\documentclass{article}

\usepackage{PRIMEarxiv}

\usepackage[utf8]{inputenc} 
\usepackage[T1]{fontenc}    
\usepackage{hyperref}       
\usepackage{url}            
\usepackage{booktabs}       
\usepackage{amsfonts}       
\usepackage{amsmath}
\usepackage{amssymb}
\usepackage{nicefrac}       
\usepackage{microtype}      
\usepackage{fancyhdr}       
\usepackage{graphicx}       
\usepackage{multirow}
\usepackage{algorithm}
\usepackage{algpseudocode}
\usepackage{placeins}

\DeclareMathOperator{\atantwo}{atan2}
\DeclareMathOperator{\fftshift}{fftshift}

\title{PhaseCLIP: Phase-Aware Frequency-Domain Adaptation of Vision-Language Models for Few-Shot Remote-Sensing Classification
}

\author{
  Mehmet Sergen Catal \\
  Department of Electrical and Electronics Engineering  \\
  Izmir Institute of Technology \\
  İzmir, Türkiye \\
  \texttt{mehmetsergencatal@iyte.edu.tr} \\
  \And
  Abdurrahman Gumus \\
  Department of Computer Engineering \\
  Isparta University of Applied Sciences \\
  Isparta, Türkiye \\
  \texttt{abdurrahmangumus@isparta.edu.tr} \\
  \And
  Mehmet Zübeyir Ünlü \\
  Department of Electrical and Electronics Engineering  \\
  Izmir Institute of Technology \\
  İzmir, Türkiye \\
  \texttt{zubeyirunlu@iyte.edu.tr} \\
}

\begin{document}
\maketitle

\begin{abstract}
Remote-sensing (RS) image scene classification aims to assign a semantic category to an overhead image, and in practice must often do so from only a handful of labelled examples, since annotation requires expert interpretation and the resulting labels transfer poorly across sensors and regions. Recent progress has come from adapting vision-language models (VLMs) such as Contrastive Language–Image Pre-training (CLIP), through learned prompts or feature adapters. Although effective, these methods construct their representation entirely in the spatial domain. RS scenes are dominated by periodic texture and regular geometry, which the Fourier spectrum describes compactly through an amplitude component, carrying texture and contrast statistics, and a phase component, carrying spatial structure. Existing frequency-domain adaptation of VLMs either exploits amplitude alone or does not separate amplitude from phase, leaving the structural component unused. In this article, we propose PhaseCLIP, which inserts phase-capable spectral modules and a feature-gating module into a frozen CLIP visual encoder and trains only 1.30\% of its parameters. Because the amplitude and phase pathways share an identical parameter layout, a single weight set is learned on phase and transferred exactly to amplitude, giving a two-stage scheme that adds no parameters. Experiments on five public RS datasets, EuroSAT, PatternNet, UC Merced, LSRS and RSSCN7, against eight baselines under a matched protocol, show that PhaseCLIP attains 84.28\% mean accuracy and is the best method in 23 of 30 dataset-by-shot cells, with phase contributing 1.93 points overall and 4.36 points on the most difficult dataset. Because the modules act on the spectrum of image patches and transformer tokens and leave the text encoder untouched, the approach can be extended to other RS sensors and tasks, and to other domains in which class evidence lies in spatial structure. The source code of PhaseCLIP will be made publicly available at \url{https://github.com/miralab-ai/PhaseCLIP} upon publication of this article.
\end{abstract}

\keywords{Remote sensing classification \and Few-shot learning \and Vision-language model \and Frequency domain information \and Fourier transform \and Frequency domain adaptation}

\section{Introduction}
\label{sec:intro}
RS scene classification assigns a semantic label to an overhead image and is a foundational task for land-use mapping, crop monitoring, urban planning and disaster response~\cite{cheng2017rsscene}. Its progress has been driven largely by supervised deep learning, which depends on large annotated datasets that are markedly harder to produce here than in natural-image domains~\cite{cheng2020rsscene}. Overhead imagery cannot reliably be labelled by non-specialists: distinguishing an irrigated field from a permanent crop, or a commercial block from a dense residential one, requires domain expertise and often ancillary ground information. The labels that result also transfer poorly, since a dataset annotated for one sensor, spatial resolution or geographic region is frequently unsuitable for another~\cite{tuia2016domain}, so the effort must be repeated whenever operating conditions change.

New categories also appear continually as applications evolve, and each of them arrives with very few examples. A practical system must therefore be able to accommodate a class it has never seen from a handful of labelled images, rather than being retrained from scratch on a newly curated corpus. This requirement is what motivates few-shot learning in RS: given only $K$ labelled images per class, the goal is to adapt a model already trained on a large image collection, whether of overhead scenes or of general-purpose imagery, to the new categories while preserving the visual knowledge it already holds.

\subsection{Related works}
\label{sec:related}
Few-shot learning was first developed for natural images, where the dominant paradigms are metric learning and meta-learning. Matching Networks~\cite{vinyals2016matching} classify a query by attending over the labelled support set, Prototypical Networks~\cite{snell2017protonet} represent each class by the mean of its support embeddings and assign queries by nearest-prototype distance, and MAML~\cite{finn2017maml} instead learns an initialisation from which a few gradient steps suffice to fit a new task. These formulations were subsequently adapted to RS, which poses its own difficulties: high intra-class variability, small inter-class differences and large variation in object scale. RS-MetaNet~\cite{li2021rsmetanet} organises learning at the task level rather than the sample level to reduce overfitting, DLA-MatchNet~\cite{li2021dlamatchnet} uses channel and spatial attention to locate discriminative regions before matching, and SCL-MLNet~\cite{li2021sclmlnet} combines self-supervised contrastive learning with a multi-scale metric-learning branch. These methods established strong baselines for the low-label regime. Because they learn their representation from RS data alone, however, the quality of that representation remains bounded by the quantity of annotated RS imagery available, which returns to the constraint that motivates the few-shot setting in the first place.

Multimodal VLMs offer a way out of that circularity by supplying a representation learned at a scale no single-domain dataset can match. CLIP~\cite{radford2021clip} and ALIGN~\cite{jia2021align} are trained contrastively on hundreds of millions of image-text pairs to embed images and natural-language descriptions in a shared space, so a class is specified by a textual description rather than by a fixed label. This makes the label space open and enables zero-shot recognition, and it is precisely in the low-data regime that these models are most valuable: because the visual representation is already general, a downstream task supplies only the small amount of information needed to specialise it. Their pretraining corpus is nonetheless ground-level web photography, and RS images differ from it in viewpoint, scale convention and scene composition. One line of work responds by repeating the pretraining on RS data: RemoteCLIP~\cite{liu2024remoteclip} scales CLIP pretraining with RS specific image-text pairs, and RS5M together with GeoRSCLIP~\cite{zhang2024rs5m} assembles a five-million-pair corpus and fine-tunes CLIP on it. That approach changes the pretraining corpus, and is complementary to the question studied here, which is how a general-purpose model can be adapted efficiently when such a corpus is not available.

For adapting a general VLM to a domain it was not trained on, several families of lightweight fine-tuning methods have been developed, all of which keep the backbone frozen. Prompt learning replaces the hand-written class description with continuous vectors optimised on the few-shot set. CoOp~\cite{zhou2022coop} learns a single context shared across all classes, and demonstrated that a learned prompt substantially outperforms a hand-crafted template. CoCoOp~\cite{zhou2022cocoop} conditions that context on the input image, which improves generalisation to classes unseen during training. MaPLe~\cite{khattak2023maple} learns coupled prompts in both the visual and the text branch rather than in the text branch alone, allowing the two modalities to adapt together. ProDA~\cite{lu2022proda} models a distribution over prompts rather than a single point estimate, capturing the variability of plausible descriptions. PromptSRC~\cite{khattak2023promptsrc} regularises the learned prompt towards the frozen model's own behaviour to retain the general knowledge acquired during pretraining. SHIP~\cite{wang2023ship} synthesises features for classes with no available examples, and semantic-alignment prompt tuning~\cite{kuchibhotla2024semantic} aligns learned prompts with class semantics under a base-to-novel protocol. A second family leaves the prompt fixed and adapts the extracted features instead. CLIP-Adapter~\cite{gao2024clipadapter} appends a small bottleneck to the image embedding and blends its output with the original through a residual connection. Tip-Adapter~\cite{zhang2022tipadapter} builds a key-value cache directly from the few-shot images and requires no training at all, while its fine-tuned variant Tip-Adapter-F optimises the cache keys for further gains. Together these methods have made few-shot adaptation of VLMs both practical and inexpensive, and they operate on the spatial representation that the frozen encoder produces.

An image can equally be described by its spatial-frequency content: the two-dimensional Fourier transform~\cite{fourier1955heat} splits it into an amplitude spectrum, which records how much energy each spatial frequency carries, and a phase spectrum, which records how those frequency components are aligned in space. The two are not equally informative. Oppenheim and Lim~\cite{oppenheim1981phase} showed that an image reconstructed from its phase alone remains recognisable whereas one reconstructed from its amplitude alone does not, Piotrowski and Campbell~\cite{piotrowski1982phase} confirmed the asymmetry perceptually, and convolutional networks have since been shown to rely disproportionately on amplitude~\cite{chen2021apr}, so the structural content carried by phase is easily underused. The efficiency of the fast Fourier transform (FFT)~\cite{cooley1965fft} makes this spectral representation practical to compute inside a network; Section~\ref{sec:freq} examines the decomposition and its consequences for RS imagery in detail.

Deep learning has exploited this spectral representation in two rather different ways. The first treats the spectrum as a place to compute a better representation. Fast Fourier Convolution~\cite{chi2020ffc} gives convolutional layers a non-local receptive field by operating partly in the spectral domain. Global Filter Networks~\cite{rao2021gfnet} replace self-attention with learnable filters applied to the Fourier coefficients of the token grid, achieving global token mixing at log-linear cost. Adaptive Fourier Neural Operators~\cite{guibas2022afno} formulate token mixing as a continuous operator learned in the frequency domain, and SpectFormer~\cite{patro2023spectformer} shows that combining spectral layers in early blocks with attention in later ones outperforms either alone. Dynamic Spectrum Mixer~\cite{hu2023dsm} makes the spectral weights content-adaptive, adaptive frequency filters~\cite{huang2023aff} use the spectrum to build efficient global token mixers, and Wave-MLP~\cite{tang2022wavemlp} represents each token as a wave with a learned phase term that modulates how tokens are aggregated. The approach has since been carried into other domains, including facial expression recognition~\cite{tang2021fer}, multimodal fusion~\cite{lao2024spectrum}, few-shot learning~\cite{zhao2025sfif}, medical image analysis~\cite{wang2023colon,zheng2026fouriermil} and vision-language transfer~\cite{pham2025farl}.

The second line treats the spectrum as a place to compute the same representation more cheaply, exploiting the convolution theorem, which turns convolution into pointwise multiplication. Mathieu et al.~\cite{mathieu2014fft} first used FFT-based convolution to accelerate training, and the idea has been extended through tile-based decompositions~\cite{lin2019tile}, discrete cosine transform (DCT)-based spectral convolution~\cite{xu2021dct}, split convolutions~\cite{chitsaz2020split}, phasor-driven formulations~\cite{reis2024phasor} and butterfly-structured transforms~\cite{alizadeh2020butterfly}, as well as dedicated embedded and FPGA implementations~\cite{abtahi2018fft,kushchenko2022fft}; broader surveys of accelerator design cover this family in detail~\cite{dhouibi2021survey}. Frequency analysis has also been used to guide model compression by identifying which components a vision transformer (ViT) can afford to discard~\cite{wang2022vtclfc}. This line of work is concerned primarily with speed and parameter cost rather than with what the spectrum represents, but it established the computational tools on which the representational work above depends.

RS is a particularly natural setting for the representational view, because overhead scenes are dominated by periodic texture and regular geometry: the furrows of a cultivated field, the lattice of a residential block, the parallel berths of a harbour~\cite{ma1998texture}. Such content is compact in the frequency domain and diffuse in the spatial one. Fang et al.~\cite{fang2019spacefreq} showed that a joint space-frequency representation improves robustness in scene classification relative to purely location-sensitive features. LHNet~\cite{zhang2022lhnet} introduces Laplacian convolutional blocks to preserve the high-frequency detail that repeated downsampling would otherwise blur, noting that this detail is what captures within-class diversity. F3Net~\cite{shi2023f3net} applies Fourier filtering to hyperspectral classification to exchange information at long range without quadratic attention cost, and related architectures combine spatial and frequency learning for dense prediction tasks~\cite{zheng2024afidaf}.

These two lines meet in FreqDIMFT~\cite{zhang2024freqdimft}, which to our knowledge is the first method to bring frequency-domain adaptation to a VLM designed for few-shot RS. It inserts trainable spectral modules into a frozen CLIP visual encoder at two levels, on the raw image patches and inside the early transformer layers, and compresses each spectrum by a radial pooling operation that exploits its conjugate symmetry so that the parameter budget remains compatible with few-shot training. It also introduced LSRS, a large-scale benchmark used in this study. FreqDIMFT demonstrated convincingly that spectral information is a productive signal for this task. More recently, FrogDogNet~\cite{gunduboina2025frogdognet} brought Fourier filtering to CLIP prompt learning for RS domain generalisation. Building on the conditional prompts of CoCoOp, it transforms the image embedding produced by the frozen visual encoder into the frequency domain, retains its low-frequency components, and uses the filtered features to generate image-conditioned prompt tokens, improving base-to-new, cross-dataset and multi-domain generalisation on RS benchmarks. Outside RS, FE-CLIP~\cite{gong2025feclip} injects frequency information into a frozen CLIP visual encoder for zero-shot anomaly detection and segmentation of industrial and medical images. It inserts two adapters at several depths of the encoder: one applies a local DCT to windows of patch tokens, transforms the coefficients with a linear layer and returns them to the spatial domain by the inverse DCT, and the other averages sliding-window DCT responses into a map of local frequency statistics. These methods, however, draw on a single spectral quantity without distinguishing amplitude from phase: FreqDIMFT on the amplitude spectrum, FrogDogNet on low-frequency components of the global image embedding, and FE-CLIP on real-valued DCT coefficients. FE-CLIP also replaces the DCT with an FFT whose real and imaginary parts are processed separately, obtains similar accuracy, and concludes that the injection of frequency information matters more than the choice of transform; amplitude and phase are not examined separately. The phase spectrum, the component that Oppenheim and Lim identified as carrying structure, therefore remains available as a further source of information. Outside RS, the value of that component has begun to be recognised: in domain adaptation and generalisation, FDA~\cite{yang2020fda} transfers style by exchanging low-frequency amplitude while preserving phase, and FACT~\cite{xu2021fact} builds a generalisation framework on the same asymmetry.

Most directly, FARL~\cite{pham2025farl} brings the decomposition to vision-language adaptation itself. It applies the Fourier transform to the input image and then returns to the spatial domain, reconstructing a phase-only image by setting the amplitude to unity and an amplitude-only image by discarding the phase, and encoding each with a convolutional branch. Learnable representation tokens query the two resulting feature sets through parallel cross-attention blocks, and the fused tokens are injected asymmetrically into the upper layers of the text encoder while the image encoder receives the unfused tokens. FARL is evaluated on the standard base-to-novel benchmark of eleven datasets, one of which, EuroSAT, is an RS collection, and it reports substantial gains there. Its aim is to disentangle domain-invariant structure from domain-specific style so that a model generalises to unseen classes, and it accordingly treats the two spectra as parallel inputs to a fusion module that operates on reconstructed images.

\subsection{Literature gaps}
\label{sec:gaps}
Current approaches to VLM domain adaptation mostly work in the spatial domain. Prompt-learning methods act on the input to the text encoder, while adapter and cache methods act on features the frozen visual encoder has already produced. Both families are effective and inexpensive but, with few exceptions~\cite{zhang2024freqdimft,gunduboina2025frogdognet,gong2025feclip}, neither constructs a spectral representation. For imagery whose class evidence lies in periodic structure, the frequency domain therefore remains an underexploited source of complementary information.

Where the frequency domain has been used, it has been used only in part. The structural adaptation of a VLM in the frequency domain has so far relied on the amplitude spectrum~\cite{zhang2024freqdimft} or on real-valued DCT coefficients~\cite{gong2025feclip}, neither of which isolates phase, and frequency-guided prompt learning has filtered the image embedding without separating its amplitude and phase~\cite{gunduboina2025frogdognet}. The phase spectrum has been exploited for domain shift and for robustness~\cite{xu2021fact,yang2020fda}, but in convolutional networks~\cite{chen2021apr} rather than in VLMs. The one study that brings phase to a VLM reconstructs phase-only and amplitude-only images by an inverse transform and encodes them convolutionally~\cite{pham2025farl}. Its adaptation therefore acts on spatial images derived from the spectrum, not on the spectrum itself, and it targets base-to-novel generalisation on general-purpose benchmarks.

Three gaps follow. First, to our knowledge no method uses the phase spectrum by structural modification of a VLM, and none has been proposed for RS scene classification. Second, where both spectra have been used together, they have been processed in parallel; the order in which they are learned has never been examined. Third, the existing FreqDIMFT method was evaluated at a single shot count, on two datasets, and no implementation was released. Its behaviour across the shot range, and in particular in the low-shot regime where few-shot adaptation matters most, therefore remains unknown.

\subsection{Motivation and study outline}
\label{sec:motivation}
Our primary motivations for this study are:
\begin{itemize}
  \item Developing a novel phase-aware, frequency-domain adaptation of a VLM that exploits its few-shot learning capability for RS classification.
  \item Exploiting the phase spectrum, rather than amplitude alone, as the primary structural signal for adapting a VLM to overhead imagery~\cite{oppenheim1981phase}.
  \item Operating directly on the spectrum inside the visual encoder, rather than on spatially reconstructed components, so that the adaptation acts on the frequency representation itself.
  \item Determining how the amplitude and phase spectra should be combined, and whether the order in which they are learned matters.
  \item Keeping the adaptation light enough for the few-shot regime, so that exploiting a second spectrum costs no additional parameters.
\end{itemize}

In this study, we build on the module design introduced by FreqDIMFT~\cite{zhang2024freqdimft}: two spectral modules, the Local and Global Frequency Distribution Integration Modules (FDIMs), and a gating module, the Adaptive Feature Refinement Module (AFRM), inserted into a frozen CLIP visual encoder. The names FDIM and AFRM are taken from FreqDIMFT, and we retain them so that each of our modules can be traced to its original counterpart. We implement them from the algorithm given in that paper and extend them with a phase pathway, adding a variant selection block that lets each FDIM draw on the amplitude spectrum, the phase spectrum, or both, while the AFRM attenuates features made redundant by the shift from natural to overhead imagery. We then define four variants that differ only in how the two spectra are combined, ranging from the amplitude-only baseline (FreqDIMFT) to a two-stage scheme that learns phase first and refines with amplitude second. All methods are evaluated on five RS benchmarks under a single matched protocol, and we conclude with three controlled ablations isolating the contribution of phase, of data augmentation, and of a growing label space.

\subsection{Novelties and contributions}
\label{sec:contrib}
Adapting a VLM in the frequency domain gives access to information that spatial-domain methods cannot recover, and the phase spectrum in particular encodes the spatial arrangement that separates overhead scene classes. Building upon this foundation, this study presents several innovations and contributions to the field:
\begin{itemize}
  \item A novel phase-aware frequency-domain adaptation of a VLM for few-shot RS scene classification, in which the spectral modules operate on the pooled spectrum inside the visual encoder, and the text encoder remains entirely untouched.
  \item A two-stage phase-to-amplitude transfer scheme that learns the phase representation first and refines it with amplitude, reusing a single set of module weights so that no parameters are added over a single-spectrum model.
  \item A family of four variants that isolates whether the two spectra should be combined in parallel or sequentially, and in which order.
  \item An open re-implementation of the amplitude-only FreqDIMFT evaluated against seven further baselines under matched backbone, data splits, shot counts, augmentation and prompt template across five benchmarks, six shot settings and three random seeds.
  \item Demonstration of strong accuracy with only 1.30\% of the backbone parameters trainable: 84.28\% mean accuracy across five benchmarks, against 82.36\% for the amplitude-only counterpart and 79.63\% for the strongest conventional baseline, and the best result in 23 of 30 dataset-by-shot cells.
  \item A class-count scaling study shows that the phase-aware model is the most accurate method at every class count from 20 to 113 and degrades the most gracefully as the label space grows, losing 10.41 points against 11.04 for the closest baseline.
\end{itemize}

\section{Method: PhaseCLIP}
\label{sec:method}
\begin{figure}[ht]
  \centering
  \includegraphics[width=\linewidth]{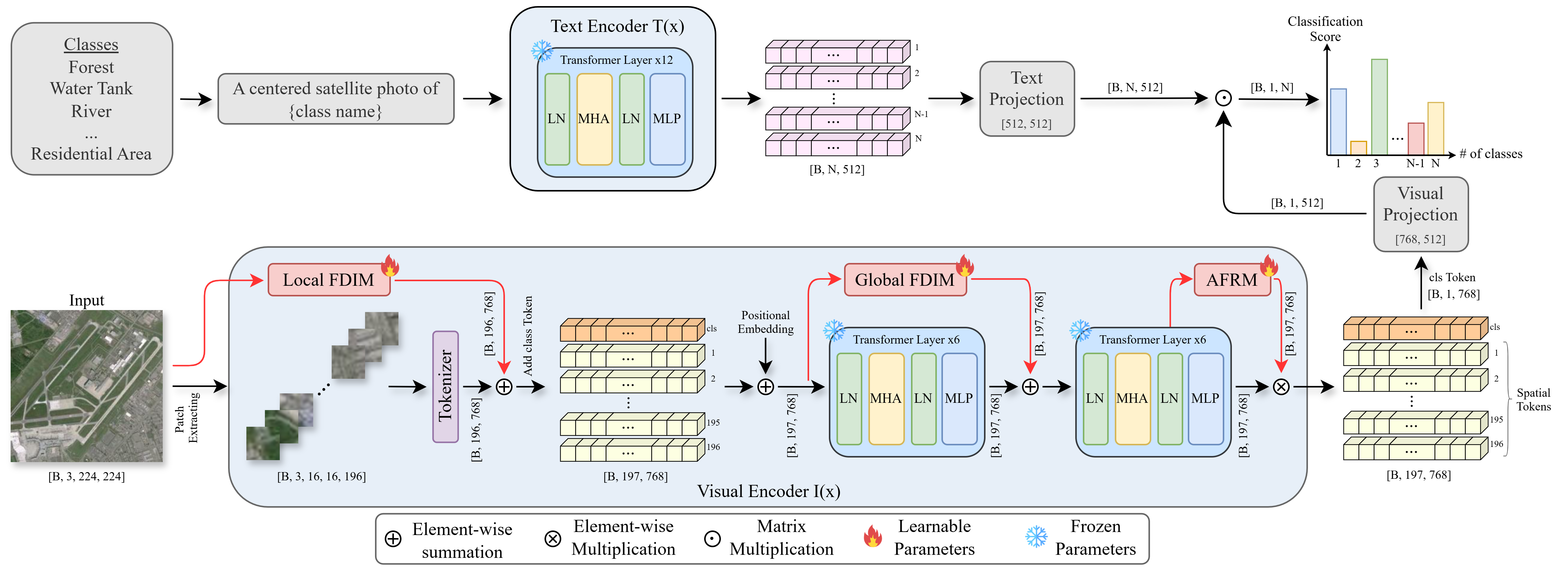}
  \caption{Overview of PhaseCLIP. A frozen CLIP dual-encoder is adapted for few-shot RS classification by three lightweight trainable modules inserted into the visual encoder: Local FDIM augments the patch embedding, Global FDIM is applied inside the first six transformer layers, and AFRM gates the last six. All CLIP weights and the entire text encoder remain frozen; an image is classified by the cosine similarity between its adapted embedding and the text prototypes of the class prompts.}
  \label{fig:framework}
\end{figure}

We adapt a frozen CLIP model to few-shot RS scene classification by inserting a small set of trainable frequency-domain modules into its visual encoder, specifically a ViT-B/16 backbone. The overall framework is shown in Figure~\ref{fig:framework}. Crucially, all CLIP parameters remain frozen; only the inserted modules are optimised, which keeps the adaptation lightweight and robust to overfitting in the low-data regime.

CLIP is a dual-encoder VLM composed of a visual encoder and a text encoder that are trained to project images and text into a shared 512-dimensional embedding space. In the ViT-B/16 visual encoder, a $224\times224$ image is divided into non-overlapping $16\times16$ patches and linearly projected into 196 patch tokens of width 768; a learnable class token is prepended and positional embeddings are added, giving 197 tokens that are processed by 12 transformer layers. The final class-token representation is layer-normalised and multiplied by a learned visual projection matrix to produce the 512-dimensional image embedding. The text encoder is a transformer that embeds a tokenised prompt, processes it with causal self-attention, and projects the representation at the end-of-text token through a learned text projection matrix into the same 512-dimensional space. Both embeddings are L2-normalised, so that their inner product equals the cosine similarity, a scale-invariant measure of the angle between the two vectors, where a larger value means the image and the text description are more closely aligned.

For classification, each class name is inserted into a single satellite-specific prompt template, ``\texttt{a centered satellite photo of \{class\}}'', which is used unchanged for every class and dataset, and encoded into a text prototype. An image is assigned to the class whose prototype has the highest cosine similarity to the image embedding, normalised across classes by a temperature-scaled softmax. In the few-shot setting, only $K$ labelled images per class are available for adaptation, with $K\in\{1,2,4,8,16,32\}$; the aim is to specialise the general-purpose CLIP representation to the RS domain from this handful of examples, without retraining the backbone. To this end we keep the entire text encoder frozen and adapt only the visual encoder, through the modules described below, which operate on the 197 transformer tokens and on the raw image patches.

\subsection{Amplitude and phase in the frequency domain}
\label{sec:freq}
Before describing the modules themselves, we make precise the two quantities they operate on. The frequency-domain view underlying PhaseCLIP rests on a single decomposition: any image or feature map can be split by the Fourier transform into an amplitude spectrum and a phase spectrum, which describe complementary aspects of its content. This subsection defines that decomposition, states what each component captures, and explains why the phase is the more valuable of the two for RS scenes. The definitions given here are used unchanged by the Local and Global FDIM modules of Section~\ref{sec:modules}.

We first fix the notation. Let $x$ be a single real-valued map, either one colour channel of an image patch or one feature channel of a token grid, of height $H$ and width $W$. Its two-dimensional discrete Fourier transform is defined in Equation~\eqref{eq:dft}.
\begin{equation}
F(u,v)=\sum_{h=0}^{H-1}\sum_{w=0}^{W-1} x(h,w)\,
       e^{-\,j\,2\pi\left(\frac{h}{H}u+\frac{w}{W}v\right)}.
\label{eq:dft}
\end{equation}
Here $(h,w)$ are the spatial coordinates of the map, with $h\in\{0,\dots,H-1\}$ indexing rows and $w\in\{0,\dots,W-1\}$ indexing columns, and $x(h,w)$ is the value of the map at that position. The output $F(u,v)$ is the complex Fourier coefficient at spatial frequency $(u,v)$, where $u$ is the vertical frequency index and $v$ the horizontal one; $j$ is the imaginary unit with $j^2=-1$, and $e^{-j2\pi(\cdot)}$ is the complex exponential that forms the basis of the transform. The transform is applied independently to every channel.

The coefficient $F(u,v)$ is complex and is fully described by its amplitude and its phase, given in Equations~\eqref{eq:amp} and~\eqref{eq:phase}.
\begin{align}
A(u,v)    &= |F(u,v)| = \sqrt{F_{\mathrm{re}}(u,v)^2 + F_{\mathrm{im}}(u,v)^2}, \label{eq:amp}\\
\varphi(u,v) &= \angle F(u,v) = \atantwo\!\big(F_{\mathrm{im}}(u,v),\,F_{\mathrm{re}}(u,v)\big), \label{eq:phase}
\end{align}
In these equations $F_{\mathrm{re}}(u,v)$ and $F_{\mathrm{im}}(u,v)$ are the real and imaginary parts of $F(u,v)$, the bars $|\cdot|$ denote the complex modulus, and $A(u,v)$ is the amplitude, or magnitude, spectrum. The symbol $\angle$ denotes the angle of a complex number, and $\atantwo$ is the two-argument arctangent, which returns that angle in the interval $[-\pi,\pi]$; $\varphi(u,v)$ is the phase spectrum. Equation~\eqref{eq:polar} writes $F$ in polar form, $A\,e^{j\varphi}$, showing that the amplitude and the phase together determine $F$ completely, so the map can be recovered from them without loss by the inverse transform. Because $x$ is real, its spectrum is conjugate-symmetric, meaning $F(-u,-v)$ equals the complex conjugate of $F(u,v)$, a property that the pooling of Section~\ref{sec:modules} exploits.
\begin{equation}
F(u,v) = A(u,v)\,e^{\,j\varphi(u,v)}.
\label{eq:polar}
\end{equation}

\begin{figure}[ht]
  \centering
  \includegraphics[width=\linewidth]{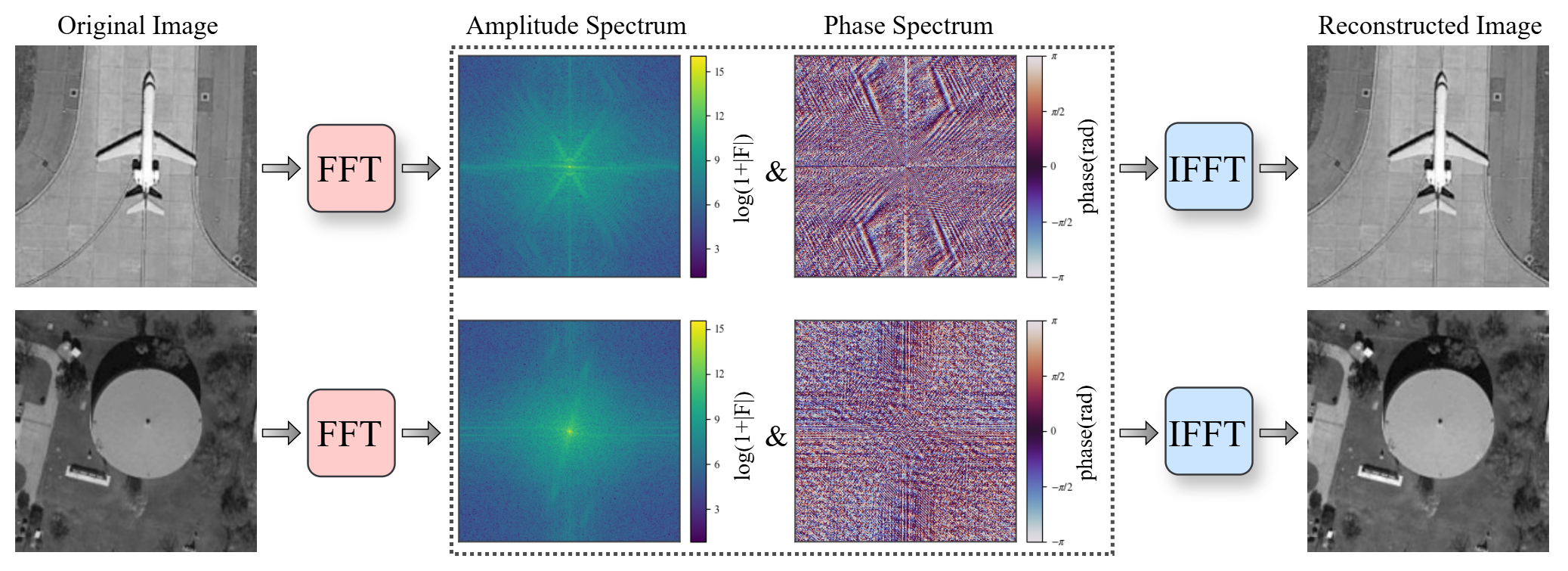}
  \caption{Fourier decomposition of an RS image. The 2-D FFT maps an image into an amplitude spectrum and a phase spectrum; the 2-D IFFT of the two together recovers the original image. The amplitude captures the energy of each spatial frequency (texture and contrast statistics), whereas the phase captures the relative alignment of those frequencies (spatial structure).}
  \label{fig:fft}
\end{figure}

Figure~\ref{fig:fft} illustrates the decomposition on an RS scene. The 2-D FFT of the image yields its amplitude spectrum $A$ and its phase spectrum $\varphi$, and the inverse FFT (IFFT) of the two together returns the original image without loss, as Equation~\eqref{eq:polar} implies. The two spectra carry very different information. The amplitude spectrum, whose energy is concentrated at the low frequencies near the centre, summarises how much texture and contrast the image contains at each spatial frequency, but says nothing about where in the image that energy lies. The phase spectrum appears to the eye as unstructured noise, yet it records the relative alignment of the frequency components, and it is this alignment that places edges, boundaries and objects at their positions in the scene.

\begin{figure}[ht]
  \centering
  \includegraphics[width=0.6\linewidth]{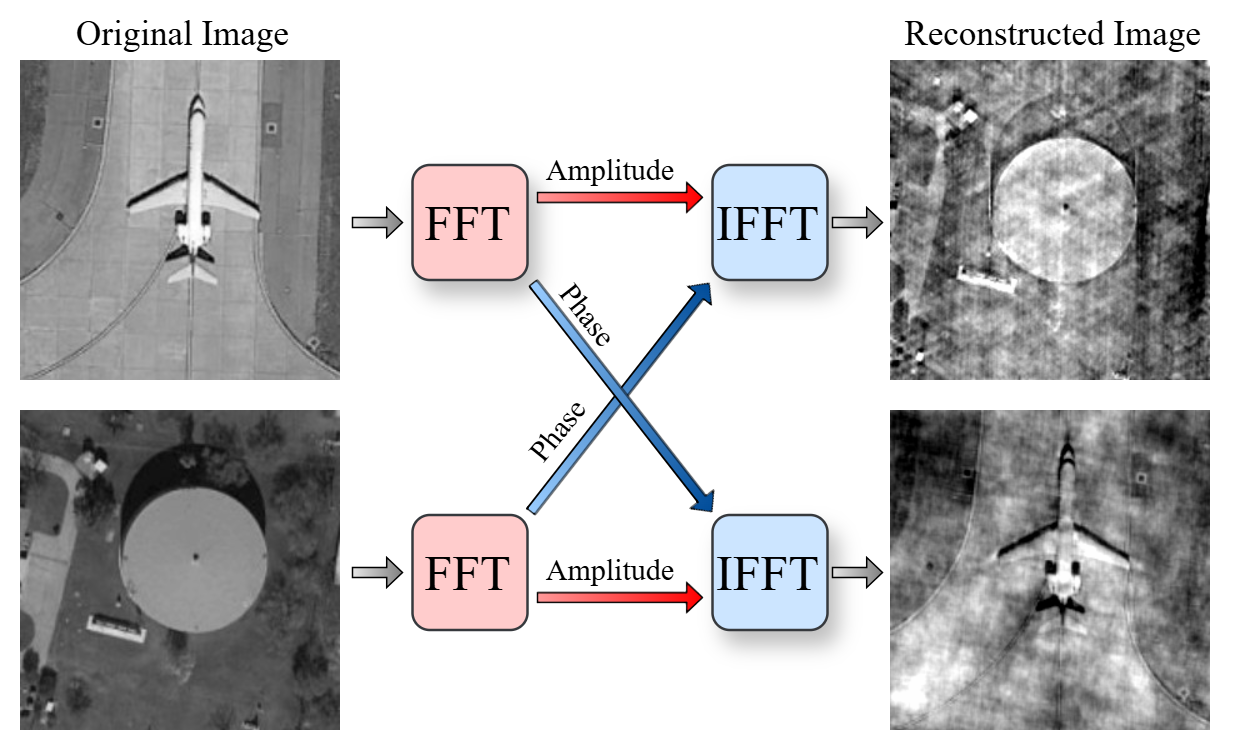}
  \caption{Phase carries the dominant structural content. Reconstructing an image from the amplitude spectrum of one scene and the phase spectrum of another yields an image that is perceived as the phase donor: the structure follows the phase while the texture energy follows the amplitude. This observation motivates the phase-aware design of PhaseCLIP. The demonstration follows Piotrowski and Campbell~\cite{piotrowski1982phase}.}
  \label{fig:phaseswap}
\end{figure}

That the two components are not equally informative is shown in Figure~\ref{fig:phaseswap}, which repeats the spectrum-exchange experiment of Piotrowski and Campbell~\cite{piotrowski1982phase} on RS scenes. An image built from the amplitude spectrum of one scene and the phase spectrum of another is perceived as the phase donor: its layout, edges and objects follow the phase, while the amplitude donor contributes only a change in overall texture and contrast. This is the perceptual counterpart of the result of Oppenheim and Lim~\cite{oppenheim1981phase}, and the same imbalance appears in learned representations, where convolutional networks depend disproportionately on amplitude and recombining the phase of one image with the amplitude of another shifts them towards the structure that phase encodes~\cite{chen2021apr}.

For RS this asymmetry is decisive, since classes such as farmland, harbour and dense residential are separated mainly by the spatial arrangement of repeated elements rather than by their average texture energy. An adaptation that relies on amplitude alone, as FreqDIMFT does, therefore leaves unused the component that best separates such classes. The modules of Section~\ref{sec:modules} can extract either spectrum, and the training scheme of Section~\ref{sec:variants} learns the phase representation first before refining it with amplitude energy, so that both the texture described by $A$ and the structure described by $\varphi$ contribute to the adapted features. As the ablation in Section~\ref{sec:abl_phase} confirms, adding the phase pathway improves accuracy on every dataset, and by the largest margin on the hardest one.

\subsection{Modules}
\label{sec:modules}
The frequency representation of Section~\ref{sec:freq} is turned into trainable features by three modules inserted into the frozen visual encoder, each acting at a different level: Local FDIM adds spectral evidence to the patch embedding, Global FDIM adds it to the feature maps of the first six transformer layers, and AFRM re-weights the representation in the last six. The two FDIMs are the spectral modules and are where the amplitude and phase information is actually used. Each takes a real map, computes its Fourier transform as in Section~\ref{sec:freq}, and keeps a spectrum chosen by a small variant selection block that can be set to amplitude only, phase only, or both in parallel. This selection is the single point of difference between the variants of Section~\ref{sec:variants}; because the amplitude and phase pathways have identical architecture and parameter shapes, a module can switch between them, or run both, without any other change, which is what makes the two-stage weight transfer of Section~\ref{sec:transfer} possible. AFRM performs no spectral operation and is shared, unchanged, across all variants. Every module uses a bottleneck of width 96, the model width 768 divided by 8, for parameter efficiency.

Both FDIMs feed a single real-valued spectrum into Distributed Average Pooling (DAP). We write this spectrum as $S$, and it is the one chosen by the module's variant selection block: either the amplitude spectrum, $S=A(u,v)$, or the phase spectrum, $S=\varphi(u,v)$, both defined in Section~\ref{sec:freq}. The complex field $F$ itself is not pooled; only the selected real spectrum $S$ is. Compression is necessary because a spectrum has as many coefficients as the map it comes from, far more than a few-shot budget can afford to fit, and for a real-valued input those coefficients are conjugate-symmetric and hence redundant. DAP exploits this symmetry by radial averaging, as defined in Equations~\eqref{eq:dap_shift} to~\eqref{eq:dap_bins}. The selected spectrum is first centred, as given in Equation~\eqref{eq:dap_shift}.
\begin{equation}
\hat{S} = \fftshift(S), \qquad S\in\{A(u,v),\varphi(u,v)\}, \qquad c = H/2.
\label{eq:dap_shift}
\end{equation}
Here $\hat{S}$ is the centred spectrum obtained by applying the Fourier shift operation $\fftshift$ to $S$, which moves the zero-frequency (direct current, DC) component to the centre index $c=H/2$. The maps are square, so $H=W$, and both sizes used here, the $16\times16$ image patch and the $14\times14$ token grid, have even side length, so $c$ is an exact integer. Only positions that lie inside the grid are pooled, and Equation~\eqref{eq:dap_valid} defines $\mathcal{V}$ as the set of all valid pixel positions of the $H\times W$ grid.
\begin{equation}
\mathcal{V} = \{0,1,\dots,H-1\}\times\{0,1,\dots,W-1\},
\label{eq:dap_valid}
\end{equation}
that is, all pairs whose row lies in $\{0,\dots,H-1\}$ and whose column lies in $\{0,\dots,W-1\}$. For each frequency offset $(u,v)$, DAP visits the four positions
\begin{equation*}
p_1=(c+u,\,c+v),\quad p_2=(c+v,\,c+u),\quad p_3=(c+u,\,c-v),\quad p_4=(c+v,\,c-u),
\end{equation*}
which are obtained from $(c+u,\,c+v)$ by reflection about the main diagonal and about the vertical axis through the centre. Their point reflections through the centre complete the set of all positions at the same frequency offset, and because the spectrum of a real map is conjugate-symmetric (Section~\ref{sec:freq}), those reflections carry no additional information and need not be visited. The positions among $p_1,\dots,p_4$ that lie inside $\mathcal{V}$ are indexed by $\Omega(u,v)$, as given in Equation~\eqref{eq:dap_omega}.
\begin{equation}
\Omega(u,v) = \bigl\{k\in\{1,2,3,4\} : p_k\in\mathcal{V}\bigr\}.
\label{eq:dap_omega}
\end{equation}
Equation~\eqref{eq:dap_mean} then defines the pooled value $D(u,v)$ as the mean of the centred spectrum $\hat{S}$ over these positions.
\begin{equation}
D(u,v) = \frac{1}{|\Omega(u,v)|}\sum_{k\in\Omega(u,v)} \hat{S}(p_k), \qquad 0\le u\le v\le H/2.
\label{eq:dap_mean}
\end{equation}
Here $|\Omega(u,v)|$ is the number of valid positions; a position that appears twice in the list, which happens when $u=0$ or $u=v$, is counted twice. For $v<H/2$ all four positions are valid, so the mean is taken over four values, whereas on the outermost ring $v=H/2$ only $p_3=(c+u,0)$ lies inside the grid, and for the corner pair $(u,v)=(H/2,H/2)$ none does. Sweeping the pair $(u,v)$ over $0\le u\le v\le H/2$ produces the number of bins $P$ given in Equation~\eqref{eq:dap_bins}.
\begin{equation}
P = \frac{\bigl(\tfrac{H}{2}+1\bigr)\bigl(\tfrac{H}{2}+2\bigr)}{2} - 1.
\label{eq:dap_bins}
\end{equation}
This gives 44 bins for a $16\times16$ image patch and 35 for the $14\times14$ token grid, a compression of about six to one; the single corner pair $(u,v)=(H/2,H/2)$ has no position inside $\mathcal{V}$ and is dropped, which is the $-1$ in Equation~\eqref{eq:dap_bins}. Applied independently to each of the $B\cdot C$ maps in a batch, DAP returns a tensor of shape $[B,C,P]$, where $B$ is the batch size, $C$ is the number of channels of the map, three colour channels in Local FDIM and 768 feature channels in Global FDIM, and $P$ is the number of bins above. We follow the DAP of the original FreqDIMFT and reimplement it from that formulation, with minor differences in the index range and in the handling of positions outside the grid, and extended from the amplitude spectrum to the selected spectrum $S$ so that it applies equally to amplitude and phase.

\subsubsection{Local FDIM}
\label{sec:local}
\begin{figure}[t]
  \centering
  \includegraphics[width=0.9\linewidth]{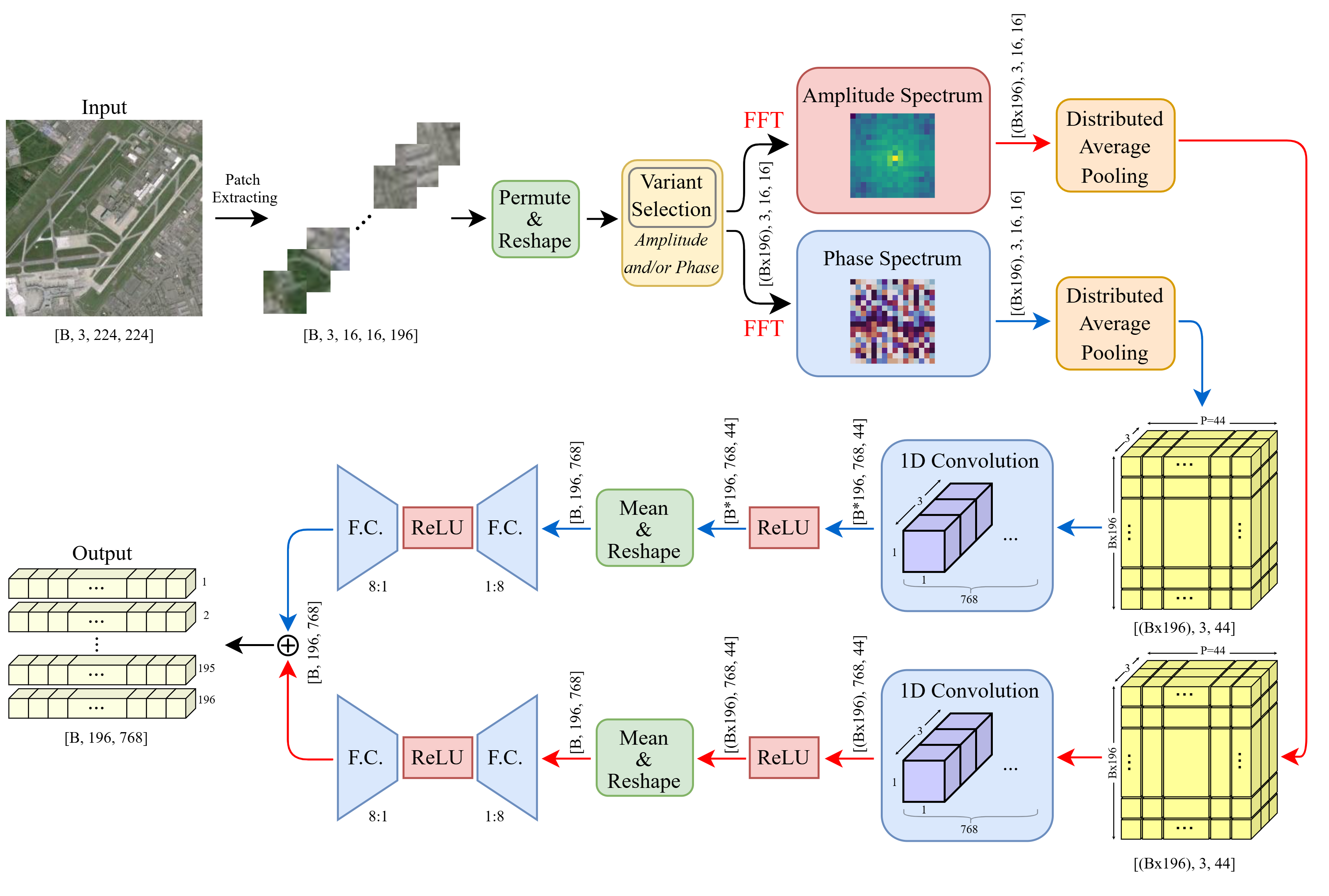}
  \caption{Local FDIM. A variant selection block chooses whether the amplitude spectrum, the phase spectrum, or both are taken from the per-patch FFT. Each $16\times16$ image patch is transformed by a 2-D FFT; the selected spectrum is compressed by DAP to 44 frequency bins, mixed across the three colour channels by a 1-D convolution and lifted to the model width ($3\to768$), averaged over the frequency bins into one token per patch, and refined by a $768\to96\to768$ bottleneck. The resulting frequency tokens are added to the convolutional patch embedding. In the parallel (both) mode, amplitude and phase are processed by two branches whose tokens are fused before the addition.}
  \label{fig:local}
\end{figure}

A $16\times16$ window of an RS scene is often a near-homogeneous texture, a crop field, a roof, a stretch of water, whose identity lies in its frequency signature rather than in its mean colour. Local FDIM, shown in Figure~\ref{fig:local}, captures that signature at the patch level, running in parallel with the standard convolutional patch embedding and producing a tensor of exactly the same shape.

Starting from the input image of shape $[B,3,224,224]$, where $B$ is the batch size and 3 is the number of colour channels, Local FDIM extracts the 196 non-overlapping $16\times16$ patches and reshapes them to $[B\cdot196,3,16,16]$, so that every patch is treated independently. A 2-D FFT is applied to each patch and, according to the variant selection block, its amplitude spectrum, its phase spectrum, or both are passed to DAP, giving $[B\cdot196,3,44]$ per selected spectrum, since $P=44$ for a $16\times16$ patch. A 1-D convolution with kernel size 1, shared across all frequency bins, mixes the three colour channels and lifts them to the model width 768, producing $[B\cdot196,768,44]$, so the module learns which colour combinations matter in each frequency band. Averaging over the 44 bins collapses this to one token per patch, $[B\cdot196,768]$, which is reshaped to $[B,196,768]$ and refined by the $768\to96\to768$ bottleneck. The resulting frequency tokens are added to the convolutional patch embeddings, before the class token and the positional embeddings are introduced. When the variant selection block is set to the both mode, used by the PhaseCLIP-SB variant, the amplitude and phase are processed by two parallel copies of this pipeline whose tokens are fused before the addition.

\subsubsection{Global FDIM}
\label{sec:global}
\begin{figure}[t]
  \centering
  \includegraphics[width=0.9\linewidth]{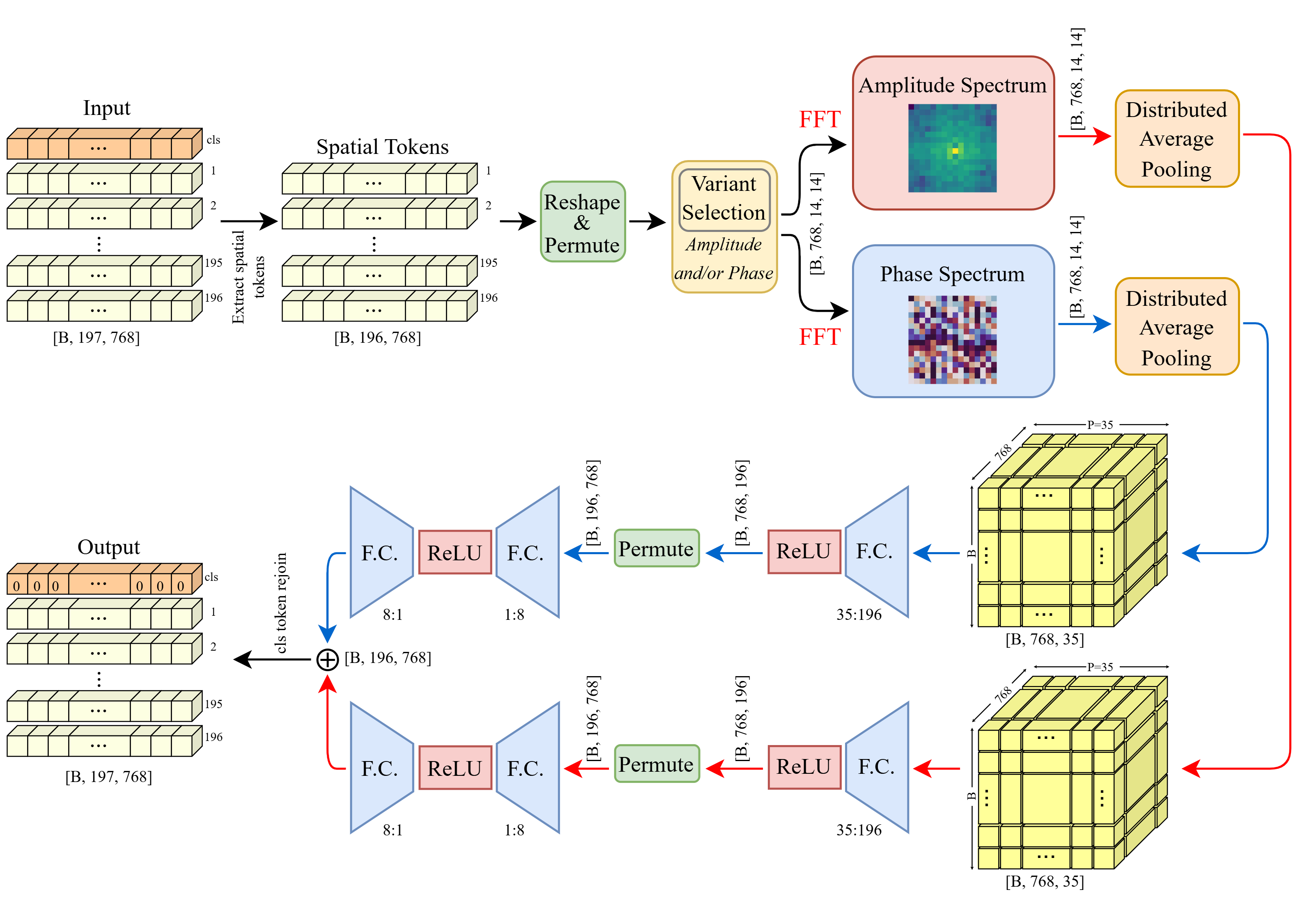}
  \caption{Global FDIM. Applied inside the first six transformer layers. A variant selection block chooses the amplitude spectrum, the phase spectrum, or both, taken from the 2-D FFT of the $14\times14$ token grid (the 768 feature channels are treated as image channels). DAP compresses the spectrum to 35 frequency bins, a learned linear layer ($35\to196$) maps these bins back to the spatial positions, and a $768\to96\to768$ bottleneck refines them; the class token receives a zero contribution and the result is added to the residual stream. In the parallel (both) mode, amplitude and phase branches are computed separately and fused before injection.}
  \label{fig:global}
\end{figure}

Global FDIM operates at the feature-map level inside the first six transformer layers, where the tokens still retain spatial layout; its structure is shown in Figure~\ref{fig:global}. It treats the 768 feature channels as the channels of a $14\times14$ image, so the FFT measures how each learned feature is distributed across the scene rather than how pixel intensities are.

From the layer features of shape $[B,197,768]$, which are the 196 spatial tokens plus one class token, each of width 768, the class token $[B,1,768]$ is set aside, since it has no spatial position and hence no meaningful spectrum, and the 196 spatial tokens are reshaped into a $14\times14$ grid and permuted to $[B,768,14,14]$. A 2-D FFT applied to the $14\times14$ grid of each of the 768 channels, followed by DAP in the mode chosen by the variant selection block, yields a descriptor of shape $[B,768,35]$, since $P=35$ for a $14\times14$ grid, which is mapped back to the spatial resolution by a learned linear projection ($35\to196$), giving $[B,768,196]$ and then $[B,196,768]$. The $768\to96\to768$ bottleneck refines these features and the class token is re-attached with a zero frequency contribution, restoring $[B,197,768]$. The module reads the layer input directly, before the layer normalisation that feeds attention, and its output is added to the residual stream after both frozen sublayers have been applied; it is thus a third parallel branch of the residual block rather than a stage interposed between them. One module is instantiated per layer, giving six independent sets of weights. As in Local FDIM, the both mode runs parallel amplitude and phase branches that are fused before injection.

\subsubsection{AFRM}
\label{sec:afrm}
\begin{figure}[t]
  \centering
  \includegraphics[width=\linewidth]{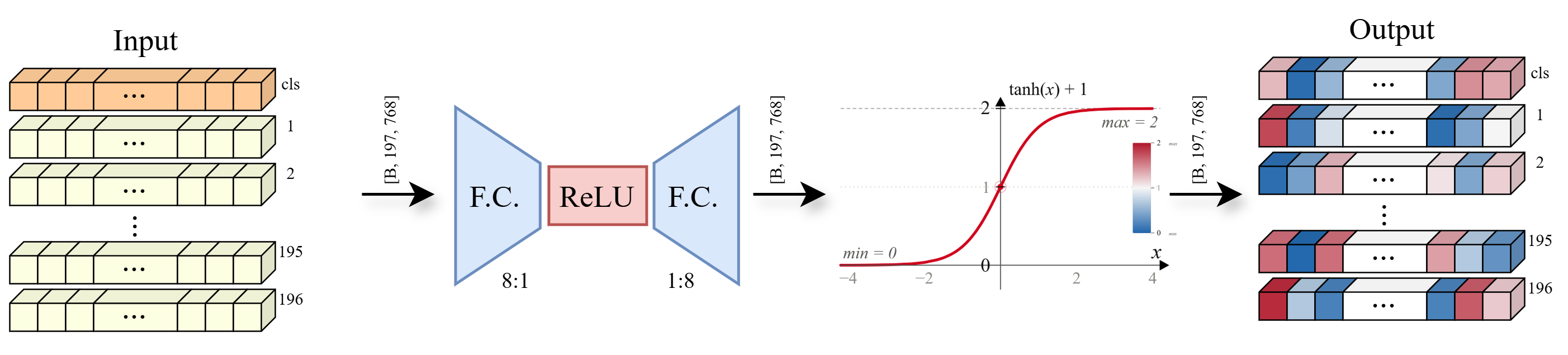}
  \caption{AFRM. Applied in the last six transformer layers, AFRM passes the $[197\times768]$ token features through a $768\to96\to768$ bottleneck and a gate $g=\tanh(\cdot)+1\in[0,2]$ that is multiplied element-wise with the features. Values below 1 suppress a feature and values above 1 amplify it, filtering out the redundancy introduced by the natural image to RS domain shift. AFRM performs no spectral operation and is shared by all variants.}
  \label{fig:afrm}
\end{figure}

CLIP is pretrained on natural images, and part of what its deep layers compute, namely features specialised to object categories and viewpoints absent from overhead imagery, is not merely unhelpful but distracting under the domain shift. A frozen backbone cannot unlearn them, so AFRM instead learns to attenuate them, in the last six transformer layers. Its structure is shown in Figure~\ref{fig:afrm}.

From the layer features $[B,197,768]$, the $768\to96$ bottleneck produces $[B,197,96]$, a rectified linear unit (ReLU) is applied, and the $96\to768$ projection restores $[B,197,768]$. A shifted activation $g=\tanh(\cdot)+1$, whose value lies in the range $[0,2]$, forms a gate of the same shape that multiplies the block output element-wise: a gate below 1 suppresses a feature, above 1 amplifies it, and near 1 leaves it unchanged. Centring the range on 1 is what distinguishes this gate from a conventional sigmoid, since the identity mapping is available in the middle of the range rather than at a saturated extreme, and because gating is per-token, the same channel may be suppressed on one token and amplified on another. AFRM takes the post-attention normalised tokens as its input and, like Global FDIM, is instantiated once per layer.

Only Local FDIM, the six Global FDIM modules and the six AFRM modules are trainable; the CLIP backbone and the text encoder are frozen throughout.

\subsection{Model variants}
\label{sec:variants}
All four variants are built from the same three modules and the same frozen backbone; they differ only in how the variant selection blocks of the two FDIMs are configured and, for the transfer variants, in the order in which the amplitude and phase representations are learned. Because the amplitude and phase pathways of an FDIM share an identical architecture and parameter layout, a single binary flag suffices to switch a module between them, and one set of weights can be trained on one spectrum and then reused on the other. We exploit this property to define a family of designs that ranges from an amplitude-only baseline to our proposed two-stage, phase-first model. Training minimises the cross-entropy classification loss on the $K$ labelled images per class.

\subsubsection{FreqDIMFT (amplitude only)}
\label{sec:var_freqdimft}
In this baseline the variant selection block of both Local FDIM and Global FDIM is fixed to the amplitude spectrum, so the model never observes phase. It is trained in a single stage. FreqDIMFT is our re-implementation of the original amplitude-based method, built on the DAP of Section~\ref{sec:modules}; it isolates the contribution of amplitude-only frequency information and serves as the reference against which the phase-aware variants are measured.

\subsubsection{PhaseCLIP-SB (separated branches)}
\label{sec:var_sb}
This variant uses amplitude and phase at the same time. The variant selection block is set to the parallel mode, so each FDIM contains two independent branches, one operating on the amplitude spectrum and one on the phase spectrum, whose output tokens are fused before they are injected into the backbone. Training is again single-stage. Because the two branches are maintained simultaneously, PhaseCLIP-SB carries a second set of frequency parameters (Section~\ref{sec:params}) and represents the most direct, but least parameter-efficient, way of combining the two spectra: the model must learn both representations jointly, with no opportunity for one to guide the other.

\subsubsection{Transfer variants: PhaseCLIP and PhaseCLIP-A2P}
\label{sec:transfer}
The two remaining variants keep a single set of module weights and learn the two spectra sequentially, transferring the weights obtained on the first spectrum as the initialisation for the second. The motivation is that a representation learned on one spectrum is not discarded when the module is switched to the other; instead it provides a strong starting point that the second stage refines. The two variants differ only in the order of the spectra: PhaseCLIP-A2P learns amplitude first and phase second, whereas our proposed PhaseCLIP learns phase first and amplitude second. The procedure for PhaseCLIP is given in Algorithm~\ref{alg:transfer}.

\begin{algorithm}[t]
\caption{Two-stage transfer training of PhaseCLIP.}
\label{alg:transfer}
\begin{algorithmic}[1]
\Statex \textbf{Stage 1 (phase).}
\State Set both FDIM variant selection blocks to the phase spectrum and initialise the Local FDIM, Global FDIM and AFRM modules randomly.
\State Train only these modules for $E$ epochs by minimising the cross-entropy classification loss, producing the phase-optimised weights $W_1$.
\Statex \textbf{Weight transfer.}
\State Switch both variant selection blocks to the amplitude spectrum and load $W_1$ into the same modules. Because the parameter layout is identical for the two spectra, the transfer is exact and adds no parameters.
\Statex \textbf{Stage 2 (amplitude).}
\State Continue training the modules, now initialised from $W_1$ rather than at random, for a further $E$ epochs on the same images, producing the final weights $W_2$.
\Statex \textbf{Output:} the adapted modules with weights $W_2$.
\end{algorithmic}
\end{algorithm}

Each stage runs for $E=100$ epochs, so a transfer variant is trained for 200 epochs in total, whereas FreqDIMFT and PhaseCLIP-SB are trained for a single 100-epoch stage. PhaseCLIP-A2P is obtained by exchanging the two spectra in Algorithm~\ref{alg:transfer}, that is, by learning amplitude in Stage 1 and phase in Stage 2.

The ordering matters because the two spectra are not equally easy to learn. As discussed in Section~\ref{sec:freq}, phase carries the spatial structure that is hardest to recover and most discriminative between scene classes, while amplitude carries texture energy that is comparatively easier to model. Learning the phase representation first forces the modules to organise around this structural signal, and the subsequent amplitude stage then refines rather than overwrites it; the reverse order, and the parallel design of PhaseCLIP-SB, both reach a weaker optimum. This is confirmed empirically in Section~\ref{sec:abl_phase}, where PhaseCLIP attains the best accuracy of the four variants on almost every dataset.

\subsection{Parameter count}
\label{sec:params}
FreqDIMFT, PhaseCLIP-A2P and PhaseCLIP each add 1,971,624 trainable parameters to the frozen 149.6M-parameter CLIP ViT-B/16, i.e.\ only 1.30\% of the 151.6M parameters of the adapted model. PhaseCLIP-SB, which maintains separate amplitude and phase branches, adds 3,053,328 parameters (2.00\%). The two-stage scheme introduces no additional parameters, since both stages reuse the same module weights.

\section{Experimental setup}
\label{sec:setup}
Because the contribution of this work rests on a like-for-like comparison, the experimental setup is specified in full. We describe the five RS benchmarks (Section~\ref{sec:datasets}), the eight baselines against which PhaseCLIP is measured (Section~\ref{sec:baselines}), the protocol held identical across every method (Section~\ref{sec:protocol}), and the optimisation settings (Section~\ref{sec:impl}).

\subsection{Datasets}
\label{sec:datasets}
\begin{figure}[t]
  \centering
  \includegraphics[width=0.85\linewidth]{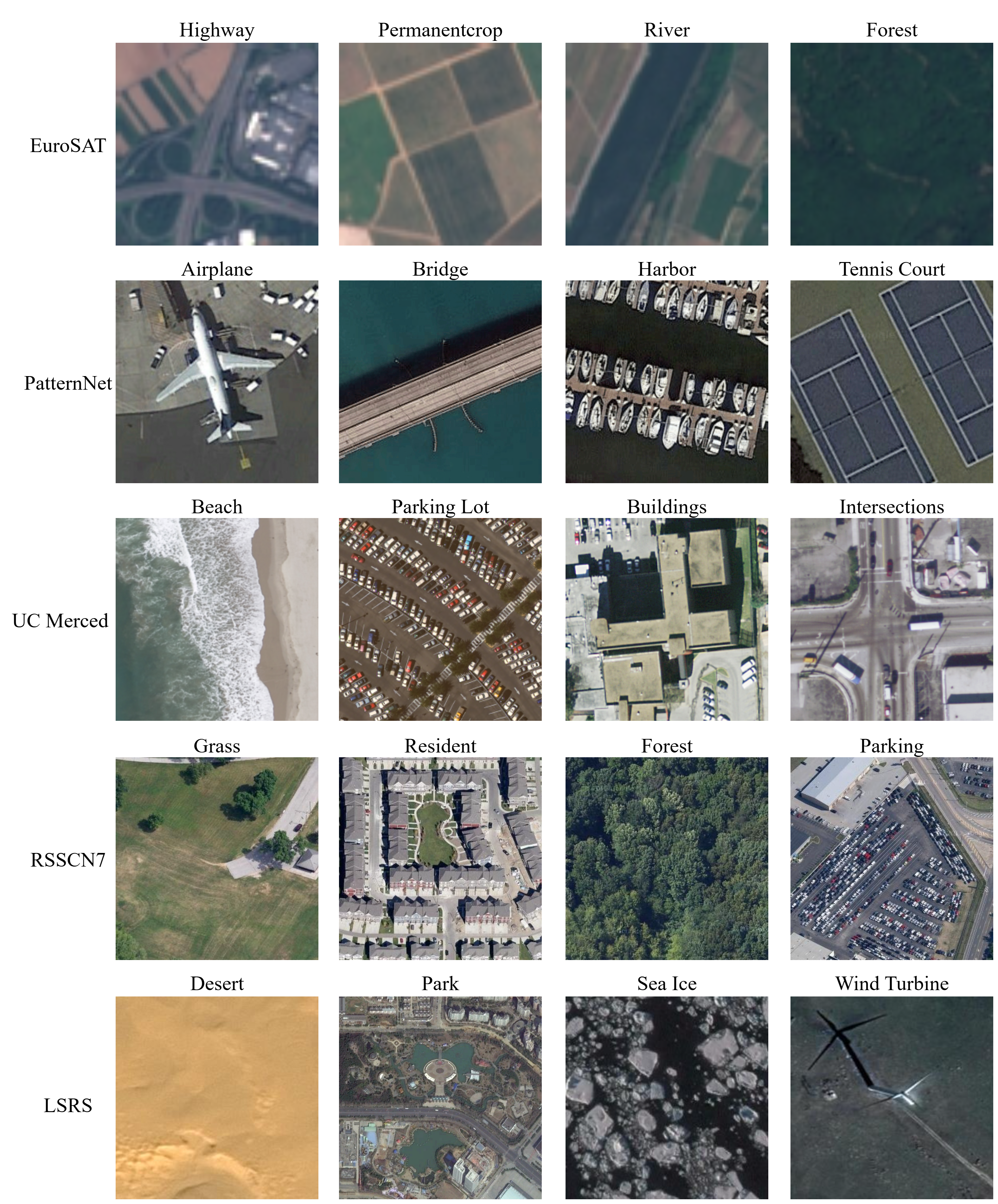}
  \caption{Representative samples from the five RS benchmarks. Each row corresponds to one dataset (EuroSAT, PatternNet, UC Merced, RSSCN7 and LSRS) and each column shows one random example scene class from that dataset. Related labels are on the top of the images.}
  \label{fig:datasets}
\end{figure}

We evaluate on five RS scene-classification benchmarks (EuroSAT~\cite{helber2019eurosat}, PatternNet~\cite{zhou2018patternnet}, UC Merced~\cite{yang2010ucmerced}, RSSCN7~\cite{zou2015rsscn7}, and LSRS~\cite{zhang2024freqdimft}), chosen to cover a broad range of image sizes, sensors, class counts and scene types; representative images from each dataset are shown in Figure~\ref{fig:datasets}. EuroSAT provides $64\times64$ multispectral tiles captured by the Sentinel-2 satellite; UC Merced consists of aerial orthoimagery from the United States Geological Survey; PatternNet and RSSCN7 are sampled from Google Earth; and LSRS is a large-scale hybrid assembled from several public RS datasets. Together they span sensors from spaceborne to aerial, image sizes from $64\times64$ to $600\times600$ pixels, and class counts from 7 to 113, so that strong performance cannot be attributed to a single sensor or scale. Their main statistics are summarised in Table~\ref{tab:datasets}.

\begin{table}[t]
\centering
\caption{Statistics of the five RS benchmarks. Each dataset is split 40\%/60\% into train/test, shared byte-identically across all methods.}
\label{tab:datasets}
\small
\setlength{\tabcolsep}{6pt}
\begin{tabular}{l c c c c c}
\toprule
Dataset & \#Classes & \#Images & \#Train & \#Test & Image size (px) \\
\midrule
EuroSAT     & 10  & 27{,}000 & 10{,}800 & 16{,}200 & $64\times64$ \\
PatternNet  & 38  & 30{,}400 & 12{,}160 & 18{,}240 & $256\times256$ \\
UC Merced   & 21  & 2{,}100  & 840      & 1{,}260  & $256\times256$ \\
RSSCN7      & 7   & 2{,}800  & 1{,}120  & 1{,}680  & $400\times400$ \\
LSRS        & 113 & 22{,}374 & 8{,}927  & 13{,}447 & $200\times200$ to $600\times600$ \\
\bottomrule
\end{tabular}
\end{table}

Among these, LSRS is a large-scale hybrid benchmark introduced by Zhang et al.~\cite{zhang2024freqdimft}, built by aggregating several publicly available RS scene datasets and merging semantically similar categories, yielding 113 scene categories of 198 images each (22,374 in total) with image sizes ranging from $200\times200$ to $600\times600$ pixels. It is the most heterogeneous and fine-grained of the five benchmarks, which makes it our primary testbed for the class-count scaling study in Section~\ref{sec:abl_classes}.

Finally, we note that semantically similar scene categories are sometimes named differently across datasets. For example, the parking-area class appears as ``Parking'' in RSSCN7 and as ``Parking Lot'' in PatternNet and UC Merced. Because every model is trained and evaluated entirely within a single dataset, using that dataset's own class names in the prompt template, these naming differences never affect any comparison.

\subsection{Baselines}
\label{sec:baselines}
We compare against eight methods covering the main families of CLIP adaptation: the untrained zero-shot CLIP~\cite{radford2021clip}; the prompt-learning methods CoOp~\cite{zhou2022coop}, CoCoOp~\cite{zhou2022cocoop} and MaPLe~\cite{khattak2023maple}; the adapter/cache methods CLIP-Adapter~\cite{gao2024clipadapter}, Tip-Adapter~\cite{zhang2022tipadapter} and Tip-Adapter-F~\cite{zhang2022tipadapter}; and FreqDIMFT~\cite{zhang2024freqdimft}, which we re-implement from its published description. Each baseline is trained and evaluated with its publicly available implementation and the default configuration released by its authors for the ViT-B/16 backbone.

\subsection{Protocol}
\label{sec:protocol}
All methods share the same experimental setup so that the comparison is fair. We use the CLIP ViT-B/16 backbone for every method and evaluate at shots $K\in\{1,2,4,8,16,32\}$. For every dataset we use a single fixed train/test split, with 40\% of the images assigned to training and the remaining 60\% to testing. The same split is used by every method and every seed, so all models are evaluated on exactly the same test images. For a $K$-shot experiment we then sample $K$ images per class from the training portion; this sample is drawn independently for each of three random seeds, and the reported accuracy is the average of the three runs on the common test set.

The text prompt is handled according to how each method is designed. Zero-shot CLIP, the adapter and cache methods (CLIP-Adapter, Tip-Adapter and Tip-Adapter-F), and the frequency-domain methods (FreqDIMFT and our PhaseCLIP) all rely on a hand-written prompt, and for these we use the same single template, ``\texttt{a centered satellite photo of \{class\}}'', for every class and dataset. The prompt-learning methods (CoOp, CoCoOp and MaPLe) do not use a fixed template: they learn their own prompt tokens during training, following their original formulations, so no template is imposed on them.

Because the backbone, data split, shots and augmentation are identical for all methods, and the text prompt is identical for every method that uses one, a difference in accuracy can be read as a difference between the adaptation methods themselves rather than between their experimental conditions. Optimisation hyperparameters are the one exception and remain method-specific, since each method is run at the settings its authors recommend.

\subsection{Implementation details}
\label{sec:impl}
Only the inserted frequency modules are optimised; the CLIP backbone and text encoder are frozen. We train with Stochastic Gradient Descent (SGD) using classical (heavy-ball) momentum~\cite{polyak1964momentum,sutskever2013momentum} set to 0.9 with no weight decay and a batch size of 128, as implemented in PyTorch~\cite{paszke2019pytorch}. The learning rate follows a linear warm-up~\cite{goyal2017warmup} over the first 4 epochs up to a base value of 0.005, after which it is decayed to zero with a cosine schedule~\cite{loshchilov2016sgdr}; gradients are clipped to a maximum global norm of 1.0~\cite{pascanu2013clipping} for stability. Each training stage runs for 100 epochs, so the two-stage transfer variants (PhaseCLIP and PhaseCLIP-A2P) are trained for 200 epochs in total. Data augmentation consists of a random resized crop~\cite{szegedy2015inception} and a random horizontal flip~\cite{krizhevsky2012alexnet}, applied identically to all adapted methods; test-time preprocessing is a deterministic resize and centre crop to $224\times224$. All experiments were run on a single NVIDIA RTX 4090 GPU with 24 GB VRAM. All share the backbone, input resolution, normalisation and augmentation described above, and all use SGD with a cosine schedule, but they differ in the settings their authors specify.

\section{Experimental results and discussion}
\label{sec:results}
This section presents the main comparison between PhaseCLIP and the eight baselines and discusses what it reveals about frequency-domain adaptation. All numbers are three-seed average accuracies obtained under the shared protocol of Section~\ref{sec:setup}; the full per-shot results are listed in Table~\ref{tab:main}. We report only the main comparison here and defer the ablations that isolate the effect of phase, of augmentation and of the number of classes to Section~\ref{sec:ablation}. One of the eight baselines is FreqDIMFT, the amplitude-only frequency method that is the closest relative of our model, so the table can be read at two levels at once: how the frequency-domain approach as a whole compares with the prompt-learning and adapter families, and how much our phase-aware extension adds on top of it. Across the five benchmarks PhaseCLIP is the strongest method overall. It obtains the highest mean accuracy on four of the five datasets and is the single best method in 23 of the 30 dataset-by-shot cells. Averaged over all datasets and shots its accuracy is 84.28\%, ahead of its amplitude-only counterpart FreqDIMFT at 82.36\% and well above the best non-frequency baseline, Tip-Adapter-F, at 79.63\%. The two frequency methods therefore occupy the top two places overall, and among them the phase-aware model is consistently the better.

\begin{table}[t]
\centering
\caption{Classification accuracy (\%) of all methods across every dataset and few-shot setting. ViT-B/16 backbone, augmentation applied to all methods, 3-seed average. The best result in each column (per dataset) is in \textbf{bold}. \textit{Avg} is the mean over the six shot settings.}
\label{tab:main}
\small
\setlength{\tabcolsep}{5pt}
\begin{tabular}{l l cccccc c}
\toprule
& & \multicolumn{6}{c}{Few-shot number} & \\
\cmidrule(lr){3-8}
Dataset & Method & 1 & 2 & 4 & 8 & 16 & 32 & Avg \\
\midrule
\multirow{9}{*}{EuroSAT} & \multicolumn{8}{c}{zero-shot CLIP\quad47.90} \\
& CoOp & 51.70 & 66.07 & 73.83 & 79.90 & 84.70 & 86.77 & 73.83 \\
& CoCoOp & 53.80 & 57.30 & 64.50 & 69.37 & 72.90 & 73.87 & 65.29 \\
& MaPLe & 44.97 & 54.10 & 73.90 & 72.17 & 85.40 & 90.30 & 70.14 \\
& CLIP-Adapter & 57.63 & 59.13 & 59.47 & 61.50 & 69.77 & 77.80 & 64.22 \\
& Tip-Adapter & 61.42 & 65.16 & 67.19 & 70.90 & 70.23 & 70.72 & 67.60 \\
& Tip-Adapter-F & 61.64 & 73.60 & 75.26 & 80.32 & 81.68 & 84.09 & 76.10 \\
& FreqDIMFT & 76.29 & 83.50 & 86.52 & 89.14 & 91.76 & 93.70 & 86.82 \\
& PhaseCLIP (ours) & \textbf{80.11} & \textbf{85.70} & \textbf{87.69} & \textbf{90.61} & \textbf{92.73} & \textbf{94.41} & \textbf{88.54} \\
\midrule
\multirow{9}{*}{PatternNet} & \multicolumn{8}{c}{zero-shot CLIP\quad59.90} \\
& CoOp & 76.07 & 83.83 & 89.27 & 92.80 & 94.67 & 96.60 & 88.87 \\
& CoCoOp & 71.37 & 77.07 & 79.37 & 84.10 & 86.90 & 89.93 & 81.46 \\
& MaPLe & 74.83 & 78.43 & 85.97 & 90.20 & 92.93 & 94.97 & 86.22 \\
& CLIP-Adapter & 65.47 & 69.50 & 71.73 & 77.47 & 85.40 & 91.70 & 76.88 \\
& Tip-Adapter & 76.82 & 81.84 & 86.57 & 90.03 & 91.28 & 91.30 & 86.31 \\
& Tip-Adapter-F & 80.89 & 85.73 & 89.95 & 92.61 & 95.39 & 96.68 & 90.21 \\
& FreqDIMFT & 81.60 & 86.78 & 91.57 & 94.64 & 96.66 & 98.15 & 91.57 \\
& PhaseCLIP (ours) & \textbf{86.05} & \textbf{90.27} & \textbf{93.63} & \textbf{95.59} & \textbf{97.27} & \textbf{98.45} & \textbf{93.54} \\
\midrule
\multirow{9}{*}{UC Merced} & \multicolumn{8}{c}{zero-shot CLIP\quad58.70} \\
& CoOp & 78.03 & 83.30 & 87.70 & 90.30 & 93.00 & 95.03 & 87.89 \\
& CoCoOp & 75.63 & 77.13 & 82.57 & 85.80 & 91.07 & 91.97 & 84.03 \\
& MaPLe & 72.03 & 78.50 & 86.20 & 88.27 & 92.50 & 94.03 & 85.25 \\
& CLIP-Adapter & 65.70 & 66.57 & 69.67 & 77.27 & 85.67 & 91.43 & 76.05 \\
& Tip-Adapter & 69.92 & 77.20 & 78.57 & 84.31 & 87.30 & 86.30 & 80.60 \\
& Tip-Adapter-F & 77.22 & \textbf{84.97} & 88.12 & 90.64 & 93.23 & 92.22 & 87.73 \\
& FreqDIMFT & 78.07 & 82.22 & 87.22 & 91.30 & 94.18 & 97.01 & 88.33 \\
& PhaseCLIP (ours) & \textbf{79.97} & 83.94 & \textbf{89.05} & \textbf{92.04} & \textbf{95.08} & \textbf{97.22} & \textbf{89.55} \\
\midrule
\multirow{9}{*}{RSSCN7} & \multicolumn{8}{c}{zero-shot CLIP\quad64.20} \\
& CoOp & 70.07 & 78.67 & 82.90 & 84.73 & 88.13 & 89.77 & 82.38 \\
& CoCoOp & 74.13 & \textbf{81.70} & 84.13 & 86.10 & 86.83 & 89.83 & 83.79 \\
& MaPLe & 72.27 & 79.10 & 83.77 & 86.37 & 87.53 & 90.07 & 83.18 \\
& CLIP-Adapter & 70.73 & 73.70 & 73.33 & 74.30 & 78.13 & 85.77 & 75.99 \\
& Tip-Adapter & 68.07 & 73.65 & 76.55 & 77.48 & 82.40 & 83.87 & 77.00 \\
& Tip-Adapter-F & 73.47 & 76.39 & 82.12 & 82.40 & 84.84 & 83.79 & 80.50 \\
& FreqDIMFT & 79.18 & 79.96 & 86.59 & 87.80 & 89.56 & \textbf{91.93} & 85.84 \\
& PhaseCLIP (ours) & \textbf{79.70} & 81.07 & \textbf{86.67} & \textbf{87.90} & \textbf{90.00} & 91.82 & \textbf{86.19} \\
\midrule
\multirow{9}{*}{LSRS} & \multicolumn{8}{c}{zero-shot CLIP\quad28.80} \\
& CoOp & 41.70 & \textbf{51.77} & 60.93 & 68.73 & 73.97 & 78.03 & 62.52 \\
& CoCoOp & 35.57 & 37.30 & 40.87 & 45.07 & 50.47 & 58.10 & 44.56 \\
& MaPLe & 39.20 & 42.33 & 45.67 & 54.30 & 61.87 & 69.70 & 52.18 \\
& CLIP-Adapter & 33.63 & 35.30 & 38.43 & 45.67 & 58.40 & 70.17 & 46.93 \\
& Tip-Adapter & 41.13 & 47.73 & 54.61 & 61.00 & 64.50 & 66.00 & 55.83 \\
& Tip-Adapter-F & \textbf{43.55} & 48.43 & \textbf{62.42} & \textbf{70.78} & 76.17 & 80.39 & \textbf{63.62} \\
& FreqDIMFT & 38.10 & 44.65 & 53.52 & 63.95 & 73.68 & 81.55 & 59.24 \\
& PhaseCLIP (ours) & 42.29 & 49.97 & 59.33 & 68.85 & \textbf{77.30} & \textbf{83.84} & 63.60 \\
\bottomrule
\end{tabular}
\end{table}

On EuroSAT the frequency-domain approach is far ahead of the conventional baselines, and phase widens the gap further. PhaseCLIP reaches 88.54\% mean accuracy and wins all six shot settings, improving on the best non-frequency baseline Tip-Adapter-F by 12.44 points; the amplitude-only FreqDIMFT, at 86.82\%, already clears every prompt-learning and adapter method by a wide margin, and PhaseCLIP adds a further 1.72 points on top of it. The advantage is largest when labels are scarcest: with a single labelled image per class PhaseCLIP attains 80.11\%, against 61.64\% for the best non-frequency baseline and 47.90\% for zero-shot CLIP. EuroSAT consists of low-resolution $64\times64$ Sentinel-2 tiles that are upsampled to the network input, so scene identity is carried almost entirely by texture and layout statistics; the size of the gap here indicates that the frequency modules capture exactly this information, and that its structural, phase-borne part is genuinely useful.

PatternNet shows the same ordering at a higher absolute level. PhaseCLIP wins all six shot settings and reaches 93.54\% mean accuracy, 3.33 points above the best non-frequency baseline Tip-Adapter-F and 1.98 points above FreqDIMFT. Because the dataset is comparatively easy, most accuracies are already high, yet PhaseCLIP still raises the 1-shot result from 80.89\% for the best non-frequency baseline to 86.05\%, and reaches 98.45\% at 32-shot. The lead is held across the entire shot range, showing that the improvement is not confined to the hardest operating point.

On UC Merced PhaseCLIP again obtains the best mean accuracy, 89.55\%, ahead of FreqDIMFT at 88.33\% and of the strongest non-frequency baseline CoOp at 87.89\%. It is the best method at five of the six shot settings; the only exception is the 2-shot column, where Tip-Adapter-F is marginally higher at 84.97\% against 83.94\%. From 4-shot onward the lead is clear and grows with supervision, reaching 97.22\% at 32-shot. UC Merced is a small set of high-resolution aerial images on which several methods are already close together, so the fact that PhaseCLIP still leads on average and at the top end confirms that frequency information stays useful even when the input is sharp and the baselines are strong.

RSSCN7 is a small seven-class benchmark on which every method performs well and the accuracies are close to saturation, and PhaseCLIP still leads on average with 86.19\%, 2.41 points above the best non-frequency baseline CoCoOp and 0.36 points above FreqDIMFT. It wins four of the six shot settings, yielding the 2-shot cell to CoCoOp and the 32-shot cell to its own amplitude-only version FreqDIMFT, each by a fraction of a point. This simply reflects that at high shots on an easy dataset the phase pathway has little left to add. Its 1-shot accuracy of 79.70\% is more than five points above the best non-frequency baseline, so the low-shot advantage seen on the other datasets holds here as well.

LSRS is by far the hardest benchmark, with 113 fine-grained classes drawn from a heterogeneous mix of source datasets, and it is the only dataset on which PhaseCLIP does not win the mean. Its average accuracy of 63.60\% is in a statistical tie with the cache-based Tip-Adapter-F at 63.62\%. What makes this dataset the most informative of the five is the position of the amplitude-only FreqDIMFT: at 59.24\% it falls to fourth place, behind not only Tip-Adapter-F but also CoOp at 62.52\%, so the frequency approach without phase is simply not competitive here. Adding the phase pathway raises the frequency method by 4.36 points, its largest gain on any dataset, and this single increment is what lifts it from fourth place to a tie for the best. The per-shot pattern explains the tie. Tip-Adapter-F is the strongest method at 1, 4 and 8 shots, and CoOp is the single best method at 2-shot, so the baselines own the low-shot end; PhaseCLIP is behind there but improves fastest as supervision grows, overtaking every method from 16-shot onward and finishing clearly ahead at 77.30\% and 83.84\% at 16 and 32 shots, against 76.17\% and 80.39\% for Tip-Adapter-F. The steep late-shot slope of its accuracy curve, together with the fact that phase alone turns a fourth-place amplitude model into a joint-best one, makes LSRS the clearest single demonstration of the value of phase information in this study, and its behaviour as the label space is varied is examined further in the class-count study of Section~\ref{sec:abl_classes}.

Taken together, these results support two conclusions. First, frequency-domain adaptation of a frozen CLIP encoder is a strong strategy for few-shot RS: the two frequency methods take the top two places overall, and PhaseCLIP is the best method on four of the five datasets, with a lead that holds across sensors, image scales and class granularities rather than being tied to a single data source. Second, the phase-aware extension is a consistent improvement over the amplitude-only design: PhaseCLIP exceeds FreqDIMFT in mean accuracy on every dataset and is clearly better on the hardest one, and it is the more robust of the two in the low-shot regime that matters most for few-shot learning. We also report the cases where other baselines remain competitive, in particular the low-shot cells of LSRS, because an honest account of them makes the overall conclusion more credible. Outside LSRS, the few cells PhaseCLIP does not win, namely the 2-shot setting on UC Merced and RSSCN7 and the 32-shot setting on RSSCN7, are lost by about one point or less and disappear as soon as the operating point moves away from those extremes. How much of the improvement is due to the phase pathway, how much to augmentation, and how the method behaves as the label space grows are the questions taken up by the ablation study in the next section.

\section{Ablation study}
\label{sec:ablation}
Having established that PhaseCLIP is the strongest method overall, we now dissect where its advantage comes from and how robust it is. This section reports three controlled ablations, each varying a single factor while holding the rest of the protocol of Section~\ref{sec:setup} fixed. Section~\ref{sec:abl_phase} isolates the contribution of the phase spectrum by comparing the amplitude-only FreqDIMFT against our three phase-aware variants, and shows both that phase carries a genuine signal and that the way it is combined with amplitude matters. Section~\ref{sec:abl_aug} measures the effect of data augmentation across all variants and datasets, confirming that it is an important ingredient in the low-shot regime and, because it is applied identically to every method, that the gains reported in Section~\ref{sec:results} are not an augmentation artefact. Section~\ref{sec:abl_classes} studies how accuracy scales as the label space grows, and shows that PhaseCLIP is not only the most accurate method at every class count but also the most robust to an increasing number of classes. Together the three studies explain the result of Section~\ref{sec:results}: the phase-first frequency design is accurate, parameter-efficient and robust, and each property can be traced to a specific design choice.

\subsection{Effect of phase: FreqDIMFT vs.\ PhaseCLIP variants}
\label{sec:abl_phase}
Table~\ref{tab:variants} isolates the contribution of the phase spectrum. It compares the amplitude-only FreqDIMFT, our re-implementation of the published method, against the three phase-aware variants introduced in Section~\ref{sec:variants} under an otherwise identical setup. Two findings stand out.

\begin{table}[t]
\centering
\caption{Accuracy comparison of the proposed PhaseCLIP variants against the amplitude-only FreqDIMFT baseline, across every dataset and few-shot setting (ViT-B/16, augmentation on, 3-seed average). Best per column (per dataset) in \textbf{bold}; \textit{Avg} is the mean over the six shots.}
\label{tab:variants}
\small
\setlength{\tabcolsep}{5pt}
\begin{tabular}{l l cccccc c}
\toprule
& & \multicolumn{6}{c}{Few-shot number} & \\
\cmidrule(lr){3-8}
Dataset & Variant & 1 & 2 & 4 & 8 & 16 & 32 & Avg \\
\midrule
\multirow{4}{*}{EuroSAT} & FreqDIMFT & 76.29 & 83.50 & 86.52 & 89.14 & 91.76 & 93.70 & 86.82 \\
 & PhaseCLIP-SB & 78.12 & 85.01 & 86.86 & 89.69 & 92.35 & 94.11 & 87.69 \\
 & PhaseCLIP-A2P & 78.57 & 85.45 & \textbf{88.58} & 90.55 & 92.39 & 94.19 & 88.29 \\
 & PhaseCLIP & \textbf{80.11} & \textbf{85.70} & 87.69 & \textbf{90.61} & \textbf{92.73} & \textbf{94.41} & \textbf{88.54} \\
\midrule
\multirow{4}{*}{PatternNet} & FreqDIMFT & 81.60 & 86.78 & 91.57 & 94.64 & 96.66 & 98.15 & 91.57 \\
 & PhaseCLIP-SB & 82.33 & 86.23 & 91.90 & 94.75 & 96.86 & 98.22 & 91.72 \\
 & PhaseCLIP-A2P & 85.64 & 90.14 & 93.50 & 95.37 & 97.23 & 98.42 & 93.38 \\
 & PhaseCLIP & \textbf{86.05} & \textbf{90.27} & \textbf{93.63} & \textbf{95.59} & \textbf{97.27} & \textbf{98.45} & \textbf{93.54} \\
\midrule
\multirow{4}{*}{UC Merced} & FreqDIMFT & 78.07 & 82.22 & 87.22 & 91.30 & 94.18 & 97.01 & 88.33 \\
 & PhaseCLIP-SB & 77.17 & 82.65 & 87.20 & \textbf{92.27} & \textbf{95.40} & 96.85 & 88.59 \\
 & PhaseCLIP-A2P & 79.31 & \textbf{84.18} & \textbf{89.31} & 92.22 & 95.05 & 97.09 & 89.53 \\
 & PhaseCLIP & \textbf{79.97} & 83.94 & 89.05 & 92.04 & 95.08 & \textbf{97.22} & \textbf{89.55} \\
\midrule
\multirow{4}{*}{RSSCN7} & FreqDIMFT & 79.18 & 79.96 & 86.59 & 87.80 & 89.56 & 91.93 & 85.84 \\
 & PhaseCLIP-SB & 78.69 & \textbf{81.23} & \textbf{87.38} & \textbf{88.10} & \textbf{90.52} & 91.87 & 86.30 \\
 & PhaseCLIP-A2P & \textbf{79.96} & 80.89 & 87.16 & 87.97 & 90.04 & \textbf{92.34} & \textbf{86.39} \\
 & PhaseCLIP & 79.70 & 81.07 & 86.67 & 87.90 & 90.00 & 91.82 & 86.19 \\
\midrule
\multirow{4}{*}{LSRS} & FreqDIMFT & 38.10 & 44.65 & 53.52 & 63.95 & 73.68 & 81.55 & 59.24 \\
 & PhaseCLIP-SB & 39.43 & 45.66 & 54.82 & 65.45 & 73.97 & 81.71 & 60.17 \\
 & PhaseCLIP-A2P & 41.76 & 49.89 & \textbf{59.41} & 68.76 & 76.92 & 83.67 & 63.40 \\
 & PhaseCLIP & \textbf{42.29} & \textbf{49.97} & 59.33 & \textbf{68.85} & \textbf{77.30} & \textbf{83.84} & \textbf{63.60} \\
\bottomrule
\end{tabular}
\end{table}

First, phase always helps. Every phase-aware variant improves on FreqDIMFT in mean accuracy averaged over the five datasets, rising from 82.36\% for the amplitude-only model to 82.89\% for the separated-branches design and to 84.20\% and 84.28\% for the two transfer variants. Our proposed PhaseCLIP, which learns phase first and then amplitude, is the best variant on four of the five datasets and improves on FreqDIMFT by 1.93 points on average. The gain appears on every dataset and is largest exactly where the task is hardest: on the 113-class LSRS benchmark phase adds 4.36 points, which, as shown in Section~\ref{sec:results}, is what lifts the frequency approach from fourth place to a tie for the best.

Second, how phase is combined with amplitude matters as much as its presence. The parallel separated-branches variant, which processes the two spectra in two simultaneous branches, gains only 0.53 points over FreqDIMFT on average, whereas the two sequential transfer variants gain 1.84 and 1.93 points. The transfer scheme is therefore both more effective and more parameter-efficient. Between the two transfer orders, learning phase first and refining with amplitude is the better choice: PhaseCLIP edges the reverse order on the overall mean and wins the per-dataset mean on four of five datasets, the two differing only on the near-saturated RSSCN7. This confirms the motivation of Section~\ref{sec:transfer}: because phase carries the structural content that is hardest to recover, organising the representation around it first and then adding amplitude energy reaches the best optimum.

\subsection{Effect of data augmentation}
\label{sec:abl_aug}
Table~\ref{tab:aug} reports the effect of data augmentation as the accuracy gain, defined as accuracy with augmentation minus accuracy without it, for every dataset, variant and shot count. The same augmentation is applied to FreqDIMFT and to all PhaseCLIP variants, so the study serves two purposes at once: it quantifies how much augmentation contributes, and it verifies that the fair comparison of Section~\ref{sec:results} is not driven by augmentation.

\begin{table}[t]
\centering
\caption{Effectiveness of augmentation, reported per dataset as the accuracy gain $\Delta = (\text{with aug}) - (\text{without aug})$ in percentage points (ViT-B/16, 3-seed average). For nearly every dataset and variant the gain is largest in the \textbf{low-shot} regime (largest value per row in \textbf{bold}) and shrinks as more labelled shots become available. \textit{Avg} is the mean over the six shots.}
\label{tab:aug}
\small
\setlength{\tabcolsep}{6pt}
\begin{tabular}{l l cccccc c}
\toprule
& & \multicolumn{6}{c}{Few-shot number} & \\
\cmidrule(lr){3-8}
Dataset & Variant & 1 & 2 & 4 & 8 & 16 & 32 & Avg \\
\midrule
\multirow{4}{*}{EuroSAT} & FreqDIMFT & +4.43 & \textbf{+6.35} & +1.76 & +0.49 & -0.02 & -0.08 & +2.15 \\
 & PhaseCLIP-SB & +0.78 & \textbf{+4.32} & +0.84 & +0.01 & -0.13 & +0.44 & +1.04 \\
 & PhaseCLIP-A2P & +3.30 & \textbf{+5.72} & +1.56 & +0.68 & -0.28 & -0.20 & +1.80 \\
 & PhaseCLIP & +5.42 & \textbf{+5.81} & -0.50 & +0.03 & -0.41 & +0.11 & +1.74 \\
\midrule
\multirow{4}{*}{PatternNet} & FreqDIMFT & \textbf{+3.40} & +3.06 & +2.33 & +1.01 & +1.42 & +1.04 & +2.04 \\
 & PhaseCLIP-SB & \textbf{+2.88} & +0.38 & +1.19 & +0.55 & +0.91 & +0.57 & +1.08 \\
 & PhaseCLIP-A2P & \textbf{+6.13} & +4.45 & +2.57 & +0.88 & +1.19 & +0.84 & +2.68 \\
 & PhaseCLIP & \textbf{+5.49} & +3.37 & +2.27 & +0.92 & +1.07 & +0.82 & +2.32 \\
\midrule
\multirow{4}{*}{UC Merced} & FreqDIMFT & \textbf{+3.07} & +1.93 & +1.14 & +0.67 & +0.74 & +0.69 & +1.37 \\
 & PhaseCLIP-SB & +1.35 & +1.09 & +0.40 & \textbf{+1.50} & +0.64 & -0.05 & +0.82 \\
 & PhaseCLIP-A2P & \textbf{+3.36} & +3.07 & +2.14 & +0.82 & +0.74 & +0.50 & +1.77 \\
 & PhaseCLIP & \textbf{+2.51} & +1.69 & +0.45 & -0.08 & +0.72 & +0.00 & +0.88 \\
\midrule
\multirow{4}{*}{RSSCN7} & FreqDIMFT & \textbf{+7.08} & +2.16 & +4.49 & +0.04 & -0.16 & -0.19 & +2.24 \\
 & PhaseCLIP-SB & +0.30 & +0.51 & \textbf{+3.26} & -0.13 & +0.08 & -0.09 & +0.65 \\
 & PhaseCLIP-A2P & \textbf{+6.29} & +2.54 & +3.63 & +0.25 & +0.14 & +0.30 & +2.19 \\
 & PhaseCLIP & \textbf{+4.86} & +1.74 & +3.46 & +0.86 & +0.20 & -0.56 & +1.76 \\
\midrule
\multirow{4}{*}{LSRS} & FreqDIMFT & +3.12 & +3.73 & \textbf{+5.94} & +5.08 & +5.04 & +3.68 & +4.43 \\
 & PhaseCLIP-SB & +1.01 & +1.40 & \textbf{+2.83} & +2.81 & +2.23 & +1.64 & +1.99 \\
 & PhaseCLIP-A2P & +4.79 & +6.32 & \textbf{+8.67} & +6.37 & +5.38 & +3.94 & +5.91 \\
 & PhaseCLIP & +4.23 & +5.18 & \textbf{+7.23} & +5.61 & +4.31 & +3.28 & +4.97 \\
\bottomrule
\end{tabular}
\end{table}

The gain is consistently positive and concentrated where labelled data are scarcest. Averaged over all datasets and variants it decreases monotonically with the number of shots, from 3.69 points at 1-shot to 3.24, 2.78, 1.42, 1.19 and 0.83 points at 2, 4, 8, 16 and 32 shots. Individual low-shot gains are large, reaching 7.08 points for FreqDIMFT at 1-shot on RSSCN7 and 8.67 points for PhaseCLIP-A2P at 4-shot on LSRS. At the high-shot end the gain approaches zero and is occasionally slightly negative, as expected once enough labelled examples are available for a crop-and-flip to add little. LSRS is the informative exception: as the hardest and most class-rich dataset it benefits from augmentation across the entire shot range, by 1.64 to 5.38 points even at 16 and 32 shots.

Two conclusions follow. Augmentation is a simple but important part of the few-shot recipe, above all in the 1-shot to 4-shot regime that matters most in practice. And because it is applied identically to every trained method, the advantage of PhaseCLIP in Section~\ref{sec:results} sits on top of augmentation rather than arising from it, which keeps the comparison fair and the conclusion robust.

\subsection{Robustness to the number of classes}
\label{sec:abl_classes}
Finally, Table~\ref{tab:classcount} studies how accuracy scales as the label space grows. Using LSRS at 16-shot, we form nested class subsets of 20, 40, 60, 80 and 113 classes, so that the only quantity changing between columns is the number of classes; each model is trained and tested on the first $N$ classes ($N\in\{20,40,60,80,113\}$). Robustness is summarised by the Drop, the accuracy at 20 classes minus the accuracy at 113 classes, where a smaller value means a method degrades more gracefully.

\begin{table}[!htbp]
\centering
\caption{Effect of the number of classes on classification accuracy (\%), on the LSRS dataset at 16-shot (ViT-B/16, 3-seed average). Models are trained and tested on nested class subsets ($20\!\subset\!40\!\subset\!60\!\subset\!80\!\subset\!113$). \textit{Drop} $=$ accuracy@20 $-$ accuracy@113 measures robustness to a growing label space (smaller is better). Best per column in \textbf{bold}; smallest \textit{Drop} in \textbf{bold}. zero-shot CLIP is an untrained reference.}
\label{tab:classcount}
\small
\setlength{\tabcolsep}{7pt}
\begin{tabular}{l ccccc c}
\toprule
& \multicolumn{5}{c}{Number of classes} & \\
\cmidrule(lr){2-6}
Method & 20 & 40 & 60 & 80 & 113 & Drop $\downarrow$ \\
\midrule
zero-shot CLIP & 46.10 & 36.30 & 34.30 & 31.40 & 28.80 & 17.30 \\
\midrule
CoOp & 86.93 & 82.03 & 78.60 & 76.33 & 73.97 & 12.96 \\
CoCoOp & 79.63 & 70.43 & 61.27 & 56.33 & 50.47 & 29.16 \\
MaPLe & 82.50 & 76.57 & 70.03 & 66.87 & 61.87 & 20.63 \\
CLIP-Adapter & 73.50 & 69.93 & 66.40 & 64.47 & 58.40 & 15.10 \\
Tip-Adapter & 78.38 & 72.94 & 69.38 & 67.09 & 64.50 & 13.88 \\
Tip-Adapter-F & 87.21 & 82.73 & 79.14 & 77.78 & 76.17 & 11.04 \\
FreqDIMFT & 85.59 & 81.46 & 77.74 & 75.68 & 73.68 & 11.91 \\
PhaseCLIP (ours) & \textbf{87.71} & \textbf{83.47} & \textbf{80.73} & \textbf{79.05} & \textbf{77.30} & \textbf{10.41} \\
\bottomrule
\end{tabular}
\end{table}

PhaseCLIP is the strongest method on this axis in two distinct senses. It is the single most accurate method at every class count, from 87.71\% at 20 classes to 77.30\% at 113 classes, ahead of every baseline across the whole range. And it degrades the least: its Drop of 10.41 points is the smallest of all methods, below Tip-Adapter-F at 11.04 and the amplitude-only FreqDIMFT at 11.91, while CoCoOp degrades most steeply at 29.16. The comparison with FreqDIMFT is especially telling, since the two share an identical architecture and differ only in the phase pathway: phase reduces the Drop by 1.50 points, so it improves not only absolute accuracy but also scaling behaviour.

Under a matched backbone, augmentation and splits, the prompt-learning methods behave inconsistently on this axis: CoOp remains reasonably robust, while CoCoOp and MaPLe degrade sharply. Robustness to a growing label space is therefore not a property of the prompt-learning family as such, and it is the phase-aware frequency approach that degrades the slowest while staying the most accurate throughout. PhaseCLIP has the potential to be the preferred method in terms of both absolute accuracy and the difficulty of the classification problem itself.
\FloatBarrier

\section{Conclusion}
\label{sec:conclusion}
In this article, we proposed PhaseCLIP, a phase-aware frequency-domain adaptation of CLIP for few-shot RS scene classification. Building on the frequency modules of FreqDIMFT, we made the spectral pathway able to draw on either spectrum and coupled the two through a two-stage scheme that learns phase first and refines it with amplitude. Because both pathways share an identical parameter layout, the transfer reuses one set of weights and adds no parameters, so only 1.30\% of the backbone is trained. Every method was evaluated under a single matched protocol, with the amplitude-only baseline re-implemented from its published algorithm.

PhaseCLIP attains the highest mean accuracy on four of the five benchmarks and is the best method in 23 of the 30 dataset-by-shot cells, averaging 84.28\% against 82.36\% for its amplitude-only counterpart and 79.63\% for the strongest conventional baseline. The margin is largest where supervision is scarcest, reaching 80.11\% at one shot on EuroSAT against 61.64\%. On LSRS, the only benchmark whose mean is not won outright, the outcome is a statistical tie, and PhaseCLIP overtakes every competing method from sixteen shots onward.

The ablation study isolates the source of this advantage. All three phase-aware variants exceed the amplitude-only model, the proposed ordering by 1.93 points on average and by 4.36 points on the 113-class LSRS benchmark, where phase alone lifts the frequency approach from fourth place to joint best. How the spectra are combined matters as much as whether phase is used: the parallel design gains 0.53 points where sequential transfer gains 1.93 at two thirds of the parameter cost. Augmentation, applied identically to every method, contributes a gain that falls monotonically with supervision and therefore does not account for the reported margins. As the label space grows from 20 to 113 classes, PhaseCLIP remains the most accurate method throughout and degrades the least of all. Together these results establish that the phase spectrum carries information a frozen VLM does not otherwise recover, and that the order in which the two spectra are learned determines how much of it is captured.

More broadly, these findings indicate that the structural content carried by the phase spectrum is not fully exploited by a pretrained visual encoder, yet it can be recovered by a small set of trainable spectral modules without altering the pretrained weights. Because the modules act on the Fourier spectrum of image patches and transformer tokens and leave the text encoder unchanged, the phase-first scheme can be extended to RS-specific VLMs, to other RS sensors and modalities, and to tasks beyond scene classification such as segmentation, detection and retrieval, as well as to other fields in which labelled data are scarce. Exploring these directions, together with open-set evaluation, is left to future work.


\bibliographystyle{unsrt}
\bibliography{references}

\end{document}